\documentclass[final,3p,times,twocolumn]{elsarticle}

\usepackage{amsmath,amssymb,mathtools}

\usepackage{amsthm}

\usepackage{booktabs}
\usepackage{array}
\usepackage{graphicx}
\usepackage{algorithm}
\usepackage{algpseudocode}
\usepackage{multirow}
\usepackage{tabularx}
\usepackage{subcaption}
\usepackage{url}
\usepackage{xcolor}
\usepackage[hidelinks]{hyperref}

\newtheorem{definition}{Definition}
\newtheorem{assumption}{Assumption}
\newtheorem{lemma}{Lemma}
\newtheorem{theorem}{Theorem}
\newtheorem{proposition}{Proposition}
\newtheorem{remark}{Remark}

\newcommand{\R}{\mathbb{R}}
\newcommand{\ip}[2]{\langle #1, #2\rangle}
\newcommand{\uhat}{\hat{u}}
\newcommand{\mhat}{\hat{m}}
\newcommand{\ent}{\mathrm{ent}}

\newcommand{\A}{\mathcal{A}}
\newcommand{\B}{\mathcal{B}}
\newcommand{\Wel}{\mathcal{W}}

\begin{document}

\begin{frontmatter}

\title{When Is a Cheap Fairness Edit Safe? Screening Post-Hoc Edits in
Two-Sided Matching Markets}

\author[duiet]{Moirangthem Tiken Singh\corref{cor1}}
\ead{tiken.m@dibru.ac.in}
\cortext[cor1]{Corresponding author.}

\author[duiet]{Rabinder Kumar Prasad}
\ead{rkp@dibru.ac.in}

\affiliation[duiet]{organization={Department of Computer Science and
Engineering, Dibrugarh University Institute of Engineering and Technology,
Dibrugarh University},
            city={Dibrugarh},
            state={Assam},
            country={India}}

\begin{abstract}
Two-sided matching systems, such as admissions pipelines, allocate
opportunities jointly. Reducing a protected attribute's influence usually
requires encoder retraining. We study a closed-form linear projector that
does this at inference time. For markets that stay vertically ordered before
and after the edit, we bound the mean welfare change by the edit-induced
appeal shift and an explicit remainder. A probabilistic refinement replaces
worst-case constants with a data-dependent deviation term. A pre-deployment
screen combines a verticality test with a welfare sweep and returns PASS,
KILL, or OUT-OF-SCOPE. An exact tier requires every opportunity to rank
applicants by appeal, and a graded tier admits near-vertical markets. On
$30$ held-out synthetic markets, the screen agrees with every pre-specified
regime verdict, with no observed false PASS. On $60$ exactly vertical
markets, no bound is violated, and the probabilistic bound is informative in
$69\%$ to $87\%$ of configurations. On $2700$ configurations from three
realistic markets, which do not reach the exact tier, no plug-in bound
expression is violated. A closed-form factorization stays within graded
scope in $90\%$ to $100\%$ of instances, whereas two-tower and graph neural
network encoders fall outside it in every instance. Architecture choice thus
appears to affect whether a post-hoc edit is admissible.
\end{abstract}


\end{frontmatter}

\section{Introduction}
\label{sec:intro}

Many machine learning systems make one-sided decisions, such as approving
a loan or selecting an item to display. Two-sided matching systems instead
allocate opportunities jointly across two populations. Examples include
school and university admissions, residency matching, ride-sharing
platforms, and freelance marketplaces. School choice and residency
matching commonly use deferred acceptance. Under deferred acceptance, an
agent's outcome depends on the scores of all other agents. Raising one
applicant's score can trigger a chain of reassignments that displaces
other applicants from their matches~\cite{roth2008deferred}. A small score
change can therefore produce a large and discontinuous change in the final
matching. Standard smoothness arguments do not directly apply to the
effect of a model adjustment on such a mechanism.

Consider a platform that scores applicants with a deployed model and
allocates them through such a mechanism. Automated systems that decide
access or admission to education are classified as high-risk under the
European Union's Artificial Intelligence Act~\cite{euaiact2024}. In such
settings, a regulator or an internal policy team may require a protected
attribute, such as race or gender, to exert less influence on the
matching. These requirements also change over time. In 2023, for example,
the Supreme Court of the United States held that race-conscious admissions
programs at two universities violated the Equal Protection
Clause~\cite{sffa2023}. The platform could retrain the model after each
change, but retraining is costly. Deep recommendation models are among the
most compute-intensive training workloads in large production data
centers~\cite{9407135}, and graph-based recommenders at web
scale are trained on multi-GPU hardware over graphs with billions of
nodes~\cite{10.1145/3219819.3219890}. Repeated
retraining may therefore be impractical, which motivates the following
research question.

\paragraph{Research question}
\emph{Can the dependence of a deployed matching system on a protected
attribute be reduced after deployment, without retraining, while providing a
provable bound on the loss in match quality and a pre-deployment check that
reports whether the bound applies?}

We address this question without differentiating the matching mechanism.
We treat the mechanism as a black box, run it before and after a
closed-form edit of the encoder output, and bound the resulting change in
mean welfare. Because deferred acceptance is discontinuous in the scores,
a bound of this kind requires structural assumptions on the market. The
central problem is therefore to identify a market structure under which
the edit has a bounded welfare cost, and to check that structure before
deployment.

\paragraph{Contributions}
\begin{itemize}\itemsep2pt

\item We formulate four requirements for a post-hoc fairness intervention
that leaves the encoder unchanged. A linear projection of the encoder
output satisfies three of them by construction. The fourth is a provable
bound on the loss in match quality.

\item For markets in which both sides share a ranking criterion before and
after the edit, we prove a bound on the change in match quality that
neither smooths nor approximates the matching mechanism. To our knowledge,
this is the first such guarantee for a post-hoc edit in two-sided
matching. An explicit form of the bound depends on quantities that can be
estimated before deployment from the representations and a labeled
validation set: the overlap between the removed information and agent
merit, the alignment of the removed information with the ranking
criterion, and the dispersion of merit among agents with equal appeal.

\item We derive a probabilistic refinement that replaces worst-case
quantities with a data-dependent deviation term at a specified failure
probability.

\item We combine the scope condition of the guarantees with an empirical
welfare sweep in a pre-deployment screen that returns \textsc{Pass},
\textsc{Kill}, or \textsc{Out-of-Scope} according to predefined scope,
gap-closure, and surplus-retention criteria.

\item We evaluate, on controlled and realistic markets, how the encoder
architecture affects the scope conditions and the deployment verdicts.

\end{itemize}

Section~\ref{sec:related} positions the work in the literature.
Section~\ref{sec:setting} formulates the problem and the intervention, and
Section~\ref{sec:theory} develops the theoretical analysis.
Section~\ref{sec:screen} presents the screening rule, and
Section~\ref{sec:results} reports the experiments.
Section~\ref{sec:discussion} interprets the findings and states the scope
of the results, and Section~\ref{sec:conclusion} concludes.

\section{Related work}
\label{sec:related}

Four lines of work address parts of the deployment problem considered
here: concept erasure, post-processing fairness, in-processing fair
ranking and recommendation, and fair matching mechanisms.
Table~\ref{tab:related} compares them on four properties of a deployable
intervention: operation after training, retargeting to a new fairness
objective without retraining, closed-form construction, and a bound on
welfare cost. Classical two-sided matching theory, reviewed last, supplies
the mechanism analysis rather than a competing intervention.

\begin{table*}[!t]
\caption{Properties of representative approaches. The welfare guarantee of
this work is conditional on vertical-market assumptions.}
\label{tab:related}
\centering
\footnotesize
\setlength{\tabcolsep}{8pt}
\renewcommand{\arraystretch}{1.08}
\begin{tabular}{lcccc}
\toprule
Method & Post-hoc & Retargetable & Closed-form & Welfare Bound \\
\midrule
Concept erasure & Yes & Yes & Partial & No \\
Post-processing fairness & Yes & Yes & Partial & No \\
In-processing fair ranking/recommendation & No & No & No & No \\
Fair matching mechanisms & N/A & N/A & N/A & No (mechanism only) \\
\midrule
\textbf{This work} & \textbf{Yes} & \textbf{Yes} & \textbf{Yes} & \textbf{Yes (conditional)} \\
\bottomrule
\end{tabular}
\end{table*}

\textbf{Concept erasure:}
Early methods removed a target attribute from static word embeddings by
projecting out an estimated bias direction~\cite{bolukbasi2016man}. Later
work introduced iterative nullspace projection~\cite{ravfogel2020null},
adversarial linear erasure~\cite{ravfogel2022linear}, and closed-form
projectors such as LEACE~\cite{belrose2023leace}. These methods prevent a
linear classifier from recovering a concept from the representation. They
operate post hoc and can be retargeted, and some admit closed-form
solutions. Their guarantees concern the linear predictability of the
attribute and the distortion of the representation. They do not bound the
allocation that a discontinuous mechanism computes from all scores
jointly.

The projector analyzed here has the same structure as LEACE for a binary
attribute. It removes the group mean-difference direction in whitened
coordinates and maps the result back. It differs in whitening the pooled
representations of both market sides and in allowing partial erasure. The
contribution of this work is therefore not the projector itself but the
analysis of its effect on the resulting allocation.

\textbf{Post-processing fairness:}
Post-processing methods adjust decisions or scores after training.
Examples include equalized-odds post-processing~\cite{hardt2016equality},
its calibrated variant~\cite{pleiss2017fairness}, and reject-option
classification~\cite{kamiran2012decision}. Fair ranking methods such as
FA*IR~\cite{zehlike2017fair} and exposure-constrained
rankings~\cite{singh2018fairness} similarly re-rank a given scored list.
Like concept erasure, these methods are post hoc and can be reapplied to
new targets. Their guarantees, however, concern individual predictions,
aggregate classifier performance, or a single ranked list. Deferred
acceptance processes all adjusted scores simultaneously, so a guarantee
for independent decisions does not transfer to the induced matching.

Representation neutralization~\cite{du2021fairnessrepresentationneutralization}
is the closest method in this group. It keeps the encoder fixed and
retrains only the classification head on neutralized representations, so
it also avoids retraining the encoder. It differs from the edit studied
here in two respects. It retrains the head for each fairness target, and
it addresses individual classification rather than a jointly computed
allocation.

\textbf{In-processing fair ranking and recommendation:}
In-processing methods build fairness into the training objective. Early
approaches learned representations that retained predictive information
while obscuring group membership~\cite{zemel2013learning}. Later work added
fairness regularizers to collaborative filtering~\cite{yao2017beyond},
decomposed embeddings into bias-aware and bias-free components for news
recommendation~\cite{wu2021fairrec}, and learned exposure-fair ranking
policies~\cite{singh2019policy}. These methods retrain the model when the
fairness criterion changes. Filter-based methods~\cite{wu2021fairgo} keep
the base encoder fixed but train a separate filter for each fairness
target. Neither class is retargetable without training, and neither
provides a closed-form inference-time edit. When the protected attribute
correlates with match quality, these methods also fix the
accuracy-fairness tradeoff during training rather than measure it at
inference time.

\textbf{Fair matching mechanisms:}
Fair matching methods change the allocation mechanism itself, through
quotas, capacity constraints, or fair stable-matching
procedures~\cite{kamada2015efficient,kojima2012school,hafalir2013effective}.
They impose distributional constraints on group representation directly.
This work instead keeps the deployed mechanism fixed and edits only its
inputs. The first three properties in Table~\ref{tab:related} describe
edits to the inputs of a fixed mechanism, so they are marked not
applicable for these methods. The two approaches are complementary, and a
platform could use both.

\textbf{Two-sided matching theory:}
The mechanism analysis builds on classical results on stability and
deferred acceptance~\cite{gale1962college}, their application to school
choice~\cite{abdulkadiroglu2003school}, and the broader market design
literature~\cite{roth1990two}. Stable assortative matching under vertical
preferences is central to the theoretical framework. Recent work learns
matching mechanisms that trade off stability and
strategy-proofness~\cite{ravindranath2023deeplearningtwosidedmatching},
whereas the mechanism here is given. To our knowledge, prior work has not
connected assortative matching theory to post-hoc edits of trained
representations or to welfare guarantees for such edits.

Among the methods reviewed here, none combines post-hoc operation,
retargetability, closed-form construction, and a welfare bound for the
induced matching. The present framework provides all four properties for
markets that satisfy explicit structural assumptions.

\section{Methodology}
\subsection{Problem formulation}
\label{sec:setting}

\subsubsection{Two-sided matching systems}
\label{sec:system}

A two-sided matching system contains two populations. Side
$\A$ contains applicants, such as users, students, or job seekers, while
side $\B$ contains opportunities, such as items, schools, or positions. A
trained encoder $E$ maps each agent to a representation, where $x_a=E(a)$,
$x_b=E(b)$, and $x_a,x_b\in\R^d$. A scorer assigns each pair the
compatibility score $s(a,b)=\ip{\phi(x_a)}{\psi(x_b)}$. In the
preference model studied here, side $\B$ ranks applicants by this score,
while side $\A$ ranks opportunities by their fixed quality $q_b$. An
allocation mechanism $\mathcal{M}$
maps the two sides' rankings to a matching $\mu=\mathcal{M}(s,q)$.
Throughout this paper,
$\mathcal{M}$ is deferred acceptance. Side $\A$ carries a protected
attribute $z\in\{0,1\}$ that may affect the scores and, consequently, the
allocation.

Only the ranking by side $\B$ depends on the edit. The qualities of side $\B$ are therefore fixed primitives, and the analysis
of Section~\ref{sec:theory} attributes any welfare change to the movement of
applicants along the quality ladder.

Throughout, the market is balanced, one to one, and complete:
$|\A|=|\B|=n$, every agent ranks every agent on the other side, and no
agent is left unmatched. Ties in the edited scores are broken by agent
index, which fixes the matching uniquely when a projection creates exact
ties.

\subsubsection{Design requirements}
\label{sec:desiderata}

A post-hoc intervention on a deployed matcher should meet four
deployment-oriented requirements, stated in Definition~\ref{def:desiderata}.

\begin{definition}[Post-hoc fairness control with bounded cost]
\label{def:desiderata}
An intervention on a trained matcher satisfies:
\begin{itemize}\itemsep2pt
\item[(D1)] It acts at inference time on the fixed encoder, without gradient
updates to it.
\item[(D2)] It is adjustable across fairness targets by changing few parameters,
without retraining.
\item[(D3)] It admits a closed form, recomputable cheaply when the attribute or
population changes.
\item[(D4)] It carries a bound on the match quality forfeited as a function of
the intervention's strength.
\end{itemize}
\end{definition}

Requirements D1--D3 distinguish post-hoc interventions
from in-processing approaches. In-processing methods generally retrain the
encoder when the protected attribute or fairness objective changes, which
violates D1 and D2. They also provide no closed-form update for inexpensive
recomputation, which violates D3. The projector introduced next satisfies
D1--D3 by construction (Section~\ref{sec:algorithms}). D4 instead requires a bound on the match quality forfeited as
a function of edit strength. Construction alone cannot provide this bound
because deferred acceptance is discontinuous in the scores. Without
additional structure, a small input perturbation need not produce a small
change in the matching. Establishing D4 is therefore the main theoretical
problem. Section~\ref{sec:theory} establishes it for markets with a shared
ranking criterion.

\subsubsection{Post-hoc fairness intervention}
\label{sec:intervention}

The deployed encoder already serves allocations, and the fairness
requirement may change faster than the encoder can be retrained. To meet
D1--D3, the intervention acts on the inference pipeline. It inserts one
linear transformation between the encoder and the scorer without modifying
either component. Let $W$ be
the whitening map for the pooled representations, and let $\uhat$ be the
between-group mean-difference direction for the whitened side-$\A$
representations. The transformation attenuates $\uhat$ in whitened
coordinates and maps the result back to the original coordinates. The
operator applied to side $\A$ is
\begin{equation}\label{eq:proj}
T_\gamma = W^{-1}\bigl(I-\gamma\,\uhat\uhat^\top\bigr)W,
\qquad \gamma\in[0,1],
\end{equation}
with $\gamma=0$ recovering the original system and $\gamma=1$ removing the
direction entirely. Scoring proceeds in the original coordinates the
deployed scorer uses, $s_\gamma(a,b)=\ip{\phi(T_\gamma x_a)}{\psi(x_b)}$,
and the mechanism produces $\mu_\gamma=\mathcal{M}(s_\gamma,q)$.

The analysis in Section~\ref{sec:theory} concerns this operator and uses
two of its properties. The inverse whitening map $W^{-1}$ is part of the
edit, so scoring remains in the geometry of the deployed scorer. The edit
acts only on side $\A$, and side-$\B$ qualities remain fixed primitives.

\subsubsection{Construction of the projector}
\label{sec:algorithms}

Algorithm~\ref{alg:erase} states the procedure that a deployment runs. It
first whitens the pooled representations of both sides. It then estimates
the protected direction $\uhat$ from the whitened group means. It erases
that direction at strength $\gamma$. It finally maps the result back
through the inverse whitening transform, which Remark~\ref{rem:identity}
shows is required for the edit to be the identity at $\gamma=0$.

\begin{algorithm}[ht]
\caption{One-sided erasure, the edit analyzed in
Theorems~\ref{thm:main} and~\ref{thm:prob}.}
\label{alg:erase}
\begin{algorithmic}[1]
\Require representations $\{x_a\}_{a\in\A}$, $\{x_b\}_{b\in\B}$, labels
$\{z_a\}_{a\in\A}$, strength $\gamma\in[0,1]$
\Ensure edited side-$\A$ representations $\{x'_a\}_{a\in\A}$; side $\B$
unchanged
\State $X \gets$ row-stack of $\{x_a\}\cup\{x_b\}$
\State $\Sigma \gets \mathrm{Cov}(X)$ \Comment{$d\times d$, independent of $z$ and $\gamma$}
\State $(Q_\Sigma,\mathrm{eigvals}) \gets \mathrm{EigenDecomposition}(\Sigma)$
\State floor the eigenvalues at $10^{-10}\max(\mathrm{eigvals})$
\Comment{singular-covariance policy}
\State $W \gets Q_\Sigma\,\mathrm{diag}(\mathrm{eigvals}^{-1/2})\,Q_\Sigma^\top$;\quad
$W^{-1} \gets Q_\Sigma\,\mathrm{diag}(\mathrm{eigvals}^{1/2})\,Q_\Sigma^\top$
\State $x_a^w \gets W x_a$ for all $a$
\State $v \gets \mathrm{mean}_{a:z_a=1}(x_a^w) - \mathrm{mean}_{a:z_a=0}(x_a^w)$
\If{$\|v\|=0$ or a group is empty}
  \State \Return $\{x_a\}$ \Comment{no estimable direction; edit is the identity}
\EndIf
\State $\uhat \gets v / \|v\|$
\For{each $a\in\A$}
  \State $x_a^{w\prime} \gets x_a^w - \gamma\,(x_a^w\cdot\uhat)\,\uhat$
\EndFor
\State $x'_a \gets W^{-1}x_a^{w\prime}$ for all $a$
\Comment{map back; see Remark~\ref{rem:identity}}
\State \Return $\{x'_a\}$, with $x'_b=x_b$ for all $b\in\B$
\end{algorithmic}
\end{algorithm}

The construction satisfies the first three requirements of
Definition~\ref{def:desiderata}. The projector acts on the output of the
fixed encoder at inference time and performs no gradient update, thereby
satisfying D1. At $\gamma=0$, the identity property establishes that the
edit family contains the deployed system.

\begin{remark}[$\gamma=0$ is exactly the identity]
\label{rem:identity}
At $\gamma=0$, line 13 leaves $x_a^{w\prime}=x_a^w$. Line 15 then computes
$W^{-1}Wx_a=x_a$ exactly, because $W^{-1}$ uses the same eigenvectors and the
reciprocal eigenvalues. Algorithm~\ref{alg:erase} therefore returns the original
representations unchanged at $\gamma=0$. The inverse map in line 15 is
necessary because whitening rescales every dimension to unit variance.
Returning $x_a^w$ would change the representation even at zero attenuation,
whereas mapping through $W^{-1}$ reproduces the deployed system exactly.
\end{remark}

The complexity of Algorithm~\ref{alg:erase} establishes
D3. Line 2 costs $O(nd^2)$, and line 3 costs $O(d^3)$ independently of
$n$. The dense transformations in lines 6 and 15 cost $O(nd^2)$, while
each remaining line costs $O(nd)$. The total cost is therefore
$O(nd^2+d^3)$, compared with $O(n^2)$ for one call to $\mathcal{M}$. When
$d^2\ll n$, the matching mechanism asymptotically dominates the projector.

Retargeting to a new attribute establishes D2. The
whitening transforms $W$ and $W^{-1}$ in Algorithm~\ref{alg:erase} depend
only on the representations, not on $z$, and can therefore be cached. Given
new labels $\{z'_a\}_{a\in\A}$ and cached representations $x_a^w=Wx_a$,
retargeting recomputes only the direction,
$\uhat' = v'/\|v'\|$ with
$v' = \mathrm{mean}_{a:z'_a=1}(x_a^w) - \mathrm{mean}_{a:z'_a=0}(x_a^w)$,
and reapplies lines 12--15 of
Algorithm~\ref{alg:erase} with $\uhat'$ replacing $\uhat$. The cost is
$O(nd)$ per target when the rank-one update is applied algebraically and
$O(nd^2)$ when each representation is multiplied by a dense $W^{-1}$.
Retargeting requires group labels for the new attribute.

Both the initial construction and retargeting produce a rank-one edit
along the direction $\uhat$. The guarantees of Section~\ref{sec:theory}
are proved for this rank-one edit.

\subsubsection{Theoretical scope}
\label{sec:open}

Because deferred acceptance is discontinuous in the scores
(Section~\ref{sec:desiderata}), the welfare analysis assumes a vertically
ordered market
in which both sides rank agents by a shared criterion. Under this structure,
the stable matching is assortative, and the effect of the edit can be
expressed as rank displacement. Markets with substantially idiosyncratic
preferences fall outside this theoretical scope.

\subsection{Theoretical analysis}
\label{sec:theory}

\subsubsection{Assortative markets}
\label{sec:assortative}

Let $k\in\R^d$ be a unit desirability direction and let
$n=|\A|=|\B|$. Applicant $a$ has appeal $\theta_a=\ip{x_a}{k}$, and
opportunity $b\in\B$ has quality $q_b\in\R$. The true utility of a match is
$U(a,b)=V(\theta^\ast_a,q_b)$, where the true merit $\theta^\ast_a$ may
differ from the observed appeal $\theta_a$ because of bias. The welfare of
a matching is $\Wel(\mu)=\tfrac1n\sum_a U(a,\mu(a))$.

The following four assumptions define the vertically
ordered markets considered in our analysis and provide the regularity
conditions needed for the welfare bounds. The first imposes a shared
ranking criterion. The remaining assumptions constrain the appeal
distribution, the quality distribution, and the utility function.

\begin{assumption}[Vertical market, V]
\label{as:vertical}
Every $b\in\B$ ranks $\A$ by descending $\theta$; every $a\in\A$ ranks $\B$ by
descending $q$; both orderings are strict.
\end{assumption}
\begin{assumption}[Bounded appeal density, D1$'$]
\label{as:d1}
There exists $\bar f<\infty$ such that for every $t\in\R,w>0$, the number of
agents with $\theta_a\in[t,t+w]$ is at most $\bar f\,w\,n+1$.
\end{assumption}
\begin{assumption}[Bounded quality spread, D2$'$]
\label{as:d2}
There exists $\Lambda<\infty$ such that $|q_{(r)}-q_{(s)}|\le\Lambda|r-s|/n$ for
all ranks $r,s$, where $q_{(r)}$ is the $r$-th largest quality.
\end{assumption}
\begin{assumption}[Lipschitz utility, L]
\label{as:lip}
There exists $M<\infty$ such that $|V(\theta^\ast,q)-V(\theta^\ast,q')|\le
M|q-q'|$ for every $\theta^\ast,q,q'$.
\end{assumption}

For the post-edit market, Assumption~\ref{as:vertical} is read with the
post-edit appeal $\theta'$ defined below in place of
$\theta$. Assumption~\ref{as:vertical} requires both sides to agree on a
common ranking criterion. It fails where preferences are idiosyncratic. In
residency matching, for example, programs and candidates rank each other by
criteria that do not reduce to a shared scalar.
Section~\ref{sec:disc-limits} returns to this case.

\begin{lemma}[Assortativity]
\label{lem:assort}
Under Assumption~\ref{as:vertical}, the unique stable matching pairs the rank-$r$
agent of $\A$ with the rank-$r$ agent of $\B$, for every $r$.
\end{lemma}

Applying $T_\gamma$ to side $\A$ changes applicant appeal.
Let $\theta'_a=\ip{T_\gamma x_a}{k}$ denote applicant $a$'s post-edit
appeal. Let $r(a)$ and $r'(a)$ denote the rank of $a$ in descending order
of $\{\theta_a\}$ and $\{\theta'_a\}$, respectively. We write $\mu=\mu_0$
and $\mu'=\mu_\gamma$ for the stable matchings before and after the edit.
Because $W$ and $W^{-1}$ are symmetric, the appeal shift for applicant $a$
is
\begin{equation}\label{eq:eta}
\zeta_a=\theta_a-\theta'_a
=\gamma\,\ip{Wx_a}{\uhat}\,\ip{W^{-1}\uhat}{k}.
\end{equation}
The coefficient $\ip{W^{-1}\uhat}{k}$ measures the alignment of the erased
direction with the desirability direction in the original coordinates. It
is not bounded by one, because $W^{-1}$ can stretch the erased direction.
Theorem~\ref{thm:main} uses $\zeta_a$ only through the $(1-\varepsilon)$
quantile of its magnitude, whereas Proposition~\ref{prop:explicit} depends
on the coefficient explicitly.
For $\varepsilon\in(0,1)$, let $\delta_\varepsilon$ be the $(1-\varepsilon)$
quantile of $\{|\zeta_a|\}$, and let $O=\{a:|\zeta_a|>\delta_\varepsilon\}$ be the
set of outliers, so that $|O|\le\varepsilon n$.

\begin{lemma}[Local displacement]
\label{lem:disp}
For every non-outlier $a\notin O$,
$|r'(a)-r(a)|\le\#\{a'\ne a:|\theta_{a'}-\theta_a|\le2\delta_\varepsilon\}+|O|$.
\end{lemma}

\subsubsection{Bounded welfare guarantee}
\label{sec:theorem}

Lemma~\ref{lem:disp} bounds the rank displacement of each
non-outlying agent. Theorem~\ref{thm:main} treats outliers separately using
the full quality range. The remaining assumptions convert local rank
displacement into bounded changes in partner quality and, consequently,
bounded changes in welfare.

\begin{theorem}[Bounded-stakes welfare guarantee]
\label{thm:main}
Suppose Assumption~\ref{as:vertical} holds for both the pre-edit and the
post-edit market at the specified $\gamma\in[0,1]$. Under
Assumptions~\ref{as:d1}--\ref{as:lip}, for every $\varepsilon\in(0,1)$,
\begin{equation}
\label{eq:main}
\begin{aligned}
\bigl|\Wel(\mu)-\Wel(\mu')\bigr|
&\le C\,\delta_{\varepsilon}+\mathcal{R}_{\varepsilon},\\
\text{where}\qquad
C&=4M\Lambda\bar{f},\\
\mathcal{R}_{\varepsilon}
&=2M\Lambda\varepsilon+\frac{M\Lambda}{n}.
\end{aligned}
\end{equation}
\end{theorem}

The left-hand side is the change in mean matched utility
because $\Wel$ includes the factor $1/n$. The bound neither linearizes nor
relaxes $\mathcal{M}$. It compares the exact outputs obtained by running the
mechanism once before and once after the edit. The proof combines local
rank displacement (Lemma~\ref{lem:disp}) with bounded quality spread
(Assumption~\ref{as:d2}).

Theorem~\ref{thm:main} establishes requirement D4 of
Definition~\ref{def:desiderata} for vertical markets. In
Equation~\eqref{eq:eta}, the quantities $W$, $\uhat$, and $k$ do not depend
on $\gamma$, so every $|\zeta_a|$ is proportional to $\gamma$. The quantile
therefore satisfies $\delta_\varepsilon(\gamma)=\gamma\,\delta_\varepsilon(1)$,
and the bound becomes
$C\gamma\,\delta_\varepsilon(1)+\mathcal{R}_\varepsilon$, an explicit
function of the edit strength. The guarantee holds at each $\gamma$ for
which Assumption~\ref{as:vertical} holds before and after the edit, which
is the condition that the exact tier of the screening rule checks
(Section~\ref{sec:screen}). The remainder $\mathcal{R}_\varepsilon$ does
not vanish at $\gamma=0$, where the welfare change is zero, so the bound
is conservative for small edits.

The constants $M$, $\Lambda$, and $\bar f$ are population
quantities. We estimate each with the sample analogue required by its
assumption. Specifically, $\Lambda$ is the largest adjacent-rank quality
gap multiplied by $n$, and $M$ is the largest per-agent utility-quality
slope because Assumption~\ref{as:lip} requires a supremum. We compute
$\bar f$ over all appeal windows. For each count $j$, the narrowest window
containing $j$ agents determines the lower limit
$(j-1)/(n\,w_{\min}(j))$. Taking the largest such limit ensures that the
sample estimate satisfies Assumption~\ref{as:d1} by construction.

\subsubsection{A probabilistic refinement}
\label{sec:prob-refinement}

The constant $C=4M\Lambda\bar f$ depends on $\Lambda$.
Assumption~\ref{as:d2} requires $\Lambda$ to bound every adjacent-rank gap.
In a finite sample, the largest gap therefore determines $\Lambda$, even if
the realized displacement pattern places little or no weight on that gap.
The following refinement replaces this worst-case quantity with the average
gap under one additional assumption and a failure probability $\eta$.

Let the sorted quality gaps be $g_r=q_{(r)}-q_{(r+1)}$ and let the range be
$\Delta_q=q_{(1)}-q_{(n)}$. The gaps telescope exactly, so their mean is
$\bar g=\Delta_q/(n-1)$. Define the total realized displacement
$D=\sum_a|r'(a)-r(a)|$, computed directly from the two rank assignments.

\begin{assumption}[Exchangeable gaps, EXCH]
\label{as:ind}
Conditional on the vector of crossing multiplicities
$m=(m_1,\dots,m_{n-1})$, where $m_r$ counts the agents whose rank path
covers gap position $r$, the gap vector $(g_1,\dots,g_{n-1})$ is
exchangeable: its law is invariant under relabeling of the gap positions.
\end{assumption}

By construction $\sum_r m_r=D$. The assumption constrains only the
labeling of gap positions relative to the displacement pattern, and the
proof of Theorem~\ref{thm:prob} uses no further property of the gaps. Spacings of sorted independent
Gaussian draws are not exchangeable, because their distribution depends on
rank position. Interior spacings of sorted independent uniform draws are
exchangeable marginally. If the crossing multiplicities are independent of
the quality vector, that exchangeability also holds conditional on $m$. The independence required concerns all representations that enter the
pooled whitening map, including those of side $\B$.

\begin{theorem}[Probabilistic bounded-stakes guarantee]
\label{thm:prob}
Suppose Assumption~\ref{as:vertical} holds for both the pre-edit and the
post-edit market at the specified $\gamma\in[0,1]$. Under
Assumptions~\ref{as:lip} and~\ref{as:ind}, let $n\ge3$, put $N=n-1$,
$\bar g=\Delta_q/N$, $\bar m=D/N$, and
\begin{equation}\label{eq:tailvar}
\sigma^2=\frac{1}{N-1}\sum_{r=1}^{N}(m_r-\bar m)^2
\sum_{r=1}^{N}(g_r-\bar g)^2 .
\end{equation}
Then for every $\eta\in(0,1)$, with probability at least $1-\eta$,
\begin{equation}\label{eq:prob-bound}
\bigl|\Wel(\mu)-\Wel(\mu')\bigr| \;\le\; \frac{M}{n}
\left(D\bar g + \sqrt{\frac{\sigma^2}{\eta}}\right).
\end{equation}
If $\sigma^2=0$, the bound holds with probability one.
\end{theorem}

For $n=2$ there is a single gap, and the bound holds deterministically
without the deviation term. Theorem~\ref{thm:prob} replaces an outlier-dominated
constant with a deviation term at the cost of a nonzero failure probability.
It does not use Assumptions~\ref{as:d1} and~\ref{as:d2} because $\Delta_q$
is the realized quality range. Both $D$ and $\sigma^2$ are computed from
the pre-edit and post-edit rank assignments and the quality vector. Thus,
the bound contains no fitted constant other than $M$\footnote{The proof, provided
in the supplementary material, conditions on the crossing vector and
the unordered multiset of gaps. Chebyshev's inequality then uses the exact
conditional mean and variance of a linear permutation statistic. The proof
requires no boundedness or independence beyond Assumption~\ref{as:ind}}.

\subsubsection{The role of representation entanglement}
\label{sec:entanglement}

Theorems~\ref{thm:main} and~\ref{thm:prob} bound the welfare change through
rank displacement, but they do not identify which properties of the
representations make an edit costly. A complementary diagnostic is the
overlap between the erased direction and true merit, which can be measured
before deployment.

All directions in this subsection live in whitened coordinates, since
$\uhat$ does. Let $\theta^\ast_a=\theta^\ast_0+s_m\,\ip{Wx_a}{\mhat_w}$ for a unit
merit direction $\mhat_w$ in those coordinates and a scale $s_m>0$, with
$\mhat_w$ oriented so that $\ip{\uhat}{\mhat_w}\ge0$. Define entanglement
$\ent=\ip{\uhat}{\mhat_w}\in[0,1]$, and decompose
$\uhat=\ent\,\mhat_w+\sqrt{1-\ent^2}\,\mhat_{w\perp}$ with $\mhat_{w\perp}$
a unit vector orthogonal to $\mhat_w$.

\begin{lemma}[Merit-neutral reorderings preserve welfare]
\label{lem:neutral}
If $\theta^\ast_a=\theta^\ast_{a'}$, exchanging $a,a'$'s partners leaves welfare
unchanged.
\end{lemma}

Erasing $\uhat$ changes appeal by $\zeta_a$. Only the
component that reorders agents across merit levels can affect welfare. Even
a shift orthogonal to $\mhat_w$ can reorder agents with different merits
when merit varies among agents with similar appeal. Entanglement alone
therefore does not determine the welfare-relevant perturbation. It can be
estimated from trained representations and a labeled validation set before
editing, so it serves as a pre-deployment diagnostic. Conditional cost bounds
additionally require the appeal alignment, representation spread, and merit
regularity introduced below.

\subsubsection{An explicit residual bound}
\label{sec:residual}

Three further assumptions give an explicit inequality in terms of the
quantities of Section~\ref{sec:entanglement}. The first is the quantitative
form of Lemma~\ref{lem:neutral}.

\begin{assumption}[Bounded exchange cost, $L_{\mathrm{mix}}$]
\label{as:mix}
There exists $L_{\mathrm{mix}}<\infty$ such that
$|V(\theta_1,q)-V(\theta_1,q')-V(\theta_2,q)+V(\theta_2,q')|
\le L_{\mathrm{mix}}|\theta_1-\theta_2||q-q'|$
for all $\theta_1,\theta_2,q,q'$.
\end{assumption}
\begin{assumption}[Bounded representations, $B_x$]
\label{as:bdd}
$\|Wx_a\|_2\le B_x$ for every $a\in\A$.
\end{assumption}
\begin{assumption}[Merit regularity, $(\kappa,\nu)$]
\label{as:merit}
There exist $\kappa\ge0$ and $\nu\ge0$ such that
$|\theta^\ast_a-\theta^\ast_{a'}|\le\kappa|\theta_a-\theta_{a'}|+\nu$ for
all $a,a'\in\A$.
\end{assumption}

Setting $\theta_1=\theta_2$ in
Assumption~\ref{as:mix} yields zero and recovers the merit-neutrality result
of Lemma~\ref{lem:neutral}. The constant $\kappa$ controls how quickly merit
changes with observed appeal, while $\nu$ bounds merit dispersion among
agents with equal appeal. Both can be estimated from a labeled validation
set. Assumption~\ref{as:bdd} ensures that the spreads $A_\parallel$ and
$B_\perp$, and hence $\delta_\gamma$, are finite. The constant $B_x$ does
not appear in~\eqref{eq:explicit} because the bound is expressed through
$\delta_\gamma$.

Write $c=|\ip{W^{-1}\uhat}{k}|$ for the coefficient of
Equation~\eqref{eq:eta}, put $A_\parallel=\max_a|\ip{Wx_a}{\mhat_w}|$ and
$B_\perp=\max_a\lVert Wx_a-\ip{Wx_a}{\mhat_w}\mhat_w\rVert_2$, with
$\mhat_w$ as above, and define the shift scale
\begin{equation}\label{eq:deltagamma}
\delta_\gamma=\gamma\,c\bigl(\ent\,A_\parallel
+\sqrt{1-\ent^2}\,B_\perp\bigr)\;\ge\;\max_a|\zeta_a| .
\end{equation}
Every quantity in~\eqref{eq:deltagamma} is computable from the
representations, the labels, and $\gamma$, without running the mechanism.

\begin{proposition}[Explicit residual bound]
\label{prop:explicit}
Suppose Assumption~\ref{as:vertical} holds for both the pre-edit and the
post-edit market at the specified $\gamma\in[0,1]$. Under
Assumptions~\ref{as:d1}, \ref{as:d2}, and \ref{as:mix}--\ref{as:merit},
\begin{equation}\label{eq:explicit}
\bigl|\Wel(\mu)-\Wel(\mu')\bigr|\le
L_{\mathrm{mix}}\Lambda\bigl(2\kappa\delta_\gamma+\nu\bigr)
\Bigl(2\bar f\delta_\gamma+\frac{1}{2n}\Bigr).
\end{equation}
\end{proposition}

The proof decomposes the rank permutation into adjacent transpositions,
bounds the merit gap of each transposed pair by
$2\kappa\delta_\gamma+\nu$, and bounds their number by
Assumption~\ref{as:d1}. It is given in the supplementary material.

Expanding the right-hand side of~\eqref{eq:explicit} gives
\begin{equation}\label{eq:explicit-expanded}
L_{\mathrm{mix}}\Lambda\Bigl(4\kappa\bar f\delta_\gamma^2
+\frac{\kappa\delta_\gamma}{n}
+2\nu\bar f\delta_\gamma+\frac{\nu}{2n}\Bigr).
\end{equation}
Proposition~\ref{prop:explicit} clarifies the role of
$\ent$ in three cases. First, when $\nu=0$, the bound contains the quadratic term
$4L_{\mathrm{mix}}\Lambda\kappa\bar f\delta_\gamma^2$ plus the linear
term $L_{\mathrm{mix}}\Lambda\kappa\delta_\gamma/n$, which vanishes as $n$
grows. The quadratic order occurs because a larger edit
displaces more pairs and moves each pair across a larger merit gap. Second,
when $\nu>0$, the bound includes the term
$2L_{\mathrm{mix}}\Lambda\nu\bar f\delta_\gamma$,
which is linear in $\delta_\gamma$, and the term
$L_{\mathrm{mix}}\Lambda\nu/(2n)$, which does not vanish as
$\delta_\gamma\to0$ in this upper bound. This constant reflects the
counting bound rather than a positive welfare cost in a fixed finite market.
Strict appeal order gives a minimum separation $h>0$, and no ranks change
when $2\delta_\gamma<h$. The actual welfare change is then zero. Third,
entanglement enters only through $\delta_\gamma$, together with the appeal alignment $c$
and the spreads $A_\parallel$ and $B_\perp$. If
$\ip{W^{-1}\uhat}{k}=0$, then $\delta_\gamma=0$ and no reordering occurs,
regardless of $\ent$. If
$\ent\,A_\parallel\gg\sqrt{1-\ent^2}\,B_\perp$, then
$\delta_\gamma\approx\gamma c\,\ent\,A_\parallel$, and $\ent$ governs the
bound.

\subsection{The deployment screening rule}
\label{sec:screen}

The theory provides a formal scope condition and a
measurable cost diagnostic. Verticality is required by
Theorem~\ref{thm:main}, while Proposition~\ref{prop:explicit} identifies the
shift scale $\delta_\gamma$. This scale combines entanglement, the alignment
of the erased direction with appeal, and representation spread. The
screening rule combines a scope test with an empirical sweep to produce an
operational verdict before deployment.

\begin{definition}[Pre-deployment screening rule]
\label{def:cert}
Given a market, a target attribute, and a set of tested edit strengths
that includes $\gamma=0$, the screening rule proceeds in four steps.
\begin{itemize}\itemsep2pt
\item[(S1)] \emph{Exact scope.} The $\theta$-discordance
$\text{P2}_\theta$ is the fraction of applicant pairs $(a,a')$ for which
$\operatorname{sign}\bigl(s(a,b)-s(a',b)\bigr)
=\operatorname{sign}(\theta_a-\theta_{a'})$ fails for some $b$. It is
computed before the edit and, with $\theta'$ in place of $\theta$, at every
tested $\gamma$. $\text{P2}_\theta=0$ exactly when
Assumption~\ref{as:vertical} holds with strict orders before and after the
edit. At each $\gamma$ with $\text{P2}_\theta=0$, the market lies in the
exact tier, where Theorems~\ref{thm:main} and~\ref{thm:prob} apply.
\item[(S2)] \emph{Graded scope.} The discordance P2 is the fraction of
applicant pairs $(a,a')$ for which the sign of $s(a,b)-s(a',b)$ is not the
same for every opportunity $b$. It is computed at every tested $\gamma$.
$\text{P2}=0$ is necessary for Assumption~\ref{as:vertical} but not
sufficient, since every opportunity could agree on an order other than
descending $\theta$. The rule returns \textsc{Out-of-Scope} if the maximum
of P2 over the tested $\gamma$ exceeds a calibrated threshold
$\tau_{\mathrm{disc}}$. Otherwise the market lies in the graded tier. A
market in the exact tier at every tested $\gamma$ has $\text{P2}=0$
throughout and is therefore admitted.
\item[(S3)] \emph{Fairness and welfare sweep.} For each tested $\gamma$,
the rule computes the fraction of the group gap closed and the fraction of
assortative surplus retained, defined in
Equations~\eqref{eq:gapdef}--\eqref{eq:surpdef} below.
\item[(S4)] \emph{Verdict.} For an admitted market, the rule returns
\textsc{Pass} if some tested $\gamma$ closes at least a target fraction
$\tau_{\mathrm{gap}}$ of the gap while retaining at least a target fraction
$\tau_{\mathrm{surp}}$ of surplus, and reports the smallest such $\gamma$.
It returns \textsc{Kill} if no tested $\gamma$ meets both targets: either
every gap-closing edit drops surplus below $\tau_{\mathrm{surp}}$, or no
tested edit closes $\tau_{\mathrm{gap}}$ of the gap. Under \textsc{Kill},
it reports the smallest tested $\gamma$ that closes $\tau_{\mathrm{gap}}$
of the gap, if one exists. A zero initial group gap leaves the gap-closure
ratio undefined, and a nonpositive initial surplus leaves the retention
target undefined or uninterpretable. Either condition gives \textsc{Kill},
including when both occur together.
\end{itemize}
The entanglement $\ent$ of Section~\ref{sec:entanglement} is reported with
the verdict as an explanatory diagnostic and does not enter it.
\end{definition}

The sweep uses three quantities. Let $\mu_\gamma$ be the matching at edit
strength $\gamma$, let $q$ be the partner-quality vector, and let $z$ be
the protected attribute of side $\A$. The group gap, the welfare, and the
assortative surplus are
\begin{align}
\mathrm{gap}(\gamma)&=\Bigl|\textstyle\operatorname{mean}_{z_a=1}
q_{\mu_\gamma(a)}-\operatorname{mean}_{z_a=0}q_{\mu_\gamma(a)}\Bigr|,
\label{eq:gapdef}\\
\Wel(\mu_\gamma)&=\tfrac1n\textstyle\sum_a U\bigl(a,\mu_\gamma(a)\bigr),
\label{eq:welfaredef}\\
\mathrm{surp}(\gamma)&=\Wel(\mu_\gamma)-\overline{U},
\label{eq:surpdef}
\end{align}
where $\overline{U}$ is the mean of $U$, which is the expected welfare of a
uniformly random matching. The sweep reports the fraction of the gap
closed, $\bigl(\mathrm{gap}(0)-\mathrm{gap}(\gamma)\bigr)/\mathrm{gap}(0)$,
and the fraction of surplus retained,
$\mathrm{surp}(\gamma)/\mathrm{surp}(0)$.

Computing the gap requires only the realized matching, the
partner-quality vector, and the group labels. Computing welfare and surplus
also requires an estimate of the true utility $U$. A deployment must
estimate both $U$ and $q$, because $q$ is derived from $U$.

Decision thresholds are selected on calibration data that are independent
of the evaluation data and are held fixed during evaluation. Within the
exact tier, the guarantees of Theorems~\ref{thm:main} and~\ref{thm:prob}
apply under their remaining assumptions. Within the graded tier, a
\textsc{Pass} verdict is supported by the empirical sweep and the graded
scope criterion, not by those theorems.

\begin{remark}[Discordance and the rank-one share]
\label{rem:vertjust}
The fraction of score-matrix variance explained by the leading singular
direction, which equals $1$ when the score matrix is rank one, does not
test Assumption~\ref{as:vertical}. For a bilinear score the matrix
$\theta q^\top$ is rank one whatever the signs of $\theta$ and $q$, and a
negative entry of $q$ reverses the ranking that the corresponding agent
induces over the other side. A market in which some agents prefer the worst
partners therefore attains a leading-singular-value share of $1$ while
violating Assumption~\ref{as:vertical} for those agents. Discordance
counts a pair unless every agent on the other side orders it the same way,
so the same configuration scores near $1$.

The appeal and quality scales must therefore be oriented consistently. A
monotone positive shift preserves the rankings and priority orders that
define the market. A standardized score that assigns negative merit leaves
a rank-one score matrix intact while violating
Assumption~\ref{as:vertical}.
\end{remark}

\section{Experimental evaluation}
\label{sec:results}

\subsection{Setup}
\label{sec:setup}

\textbf{Synthetic markets:} The controlled evaluation uses nominally
vertical two-sided markets with representation
dimension $d=24$. Side-$\A$ appeal correlates with the protected attribute
at strength $\omega$, which represents merit correlation. It also contains
a component at strength $\beta$ whose sign is set by the protected
attribute and which does not enter true utility. This component represents
spurious bias. Side-$\B$ quality is drawn independently and uniformly from
a positive range. Each side-$\B$ representation scales with the quality of
its opportunity and also loads on the bias direction, allowing $\beta$ to
affect the scores. Because the pooled whitening map then depends on
quality, the crossing multiplicities need not be independent of the gaps,
and Assumption~\ref{as:ind} is not established on these markets. True utility is supermodular in appeal and quality, so
assortative sorting maximizes welfare. For the non-vertical negative
control, we replace the shared-direction structure with independent,
unstructured representations on both sides. In every experiment, the
desirability direction $k$ is the normalized mean of the side-$\B$
representations.

\textbf{Synthetic regimes:} Three synthetic regimes are used throughout. The
low-merit, high-bias regime (LM-HB) sets $\omega=0.05$ and $\beta=0.65$.
The high-merit, low-bias regime (HM-LB) sets $\omega=0.65$ and
$\beta=0.05$. The non-vertical regime (NV) is the negative control. Each
regime has a reference verdict determined by its generating parameters:
PASS for LM-HB, KILL for HM-LB, and OUT-OF-SCOPE for NV.

The exact-scope evaluation uses the same generator with two changes.
Representation noise is set to zero, and the side-$\B$ scale factors are
drawn independently of the qualities. Every side-$\B$
representation is then a positive multiple of one vector $v$, so
$k=v/\|v\|$ and every opportunity ranks applicants by $\ip{x_a}{v}$, which
is the descending $\theta$ order, before and after the one-sided edit. True utility is $U=\theta^\ast q^\top$ with $\theta^\ast>0$, so in the native coordinates
Assumption~\ref{as:lip} holds with $M=\max_a\theta^\ast_a$ and
Assumption~\ref{as:mix} with $L_{\mathrm{mix}}=1$. Assumptions~\ref{as:d1},
\ref{as:d2} and~\ref{as:merit} concern the $n$ agents of the market, so
$\bar f$, $\Lambda$ and $\nu$ are computed exactly on each instance, with
$\nu$ the maximum over all applicant pairs for the fitted $\kappa$.

\textbf{Realistic datasets:} The realistic evaluation uses three markets.
The first two are semi-synthetic: they take real applicant records and
generate the side-$\B$ quality ladder and the utility matrix. The third
uses observed ratings on both sides. The first two are score-priority
markets, in which a single merit score sets priority. The first is a
law-school-admissions-style market using LSAT score and self-reported race,
following the LSAC National Longitudinal Bar Passage
Study~\cite{wightman1998lsac}. The second is a national-exam-style market
that combines four subject scores and an essay score into one merit score,
with self-reported race or color as the protected attribute, following
Brazil's ENEM microdata~\cite{inep2023enem}. In both markets the quality of
side $\B$ is a generated monotone ladder, and true utility is the outer
product of the merit score and quality. Every encoder observes a
noise-injected interaction proxy built from this utility, as specified
below. The third market is a sparse mutual-ratings
market built from the Speed Dating experiment data~\cite{fisman2006gender}.
Each record is a directional (rater, rated, rating) triple, with the
overall liking score as the rating. Gender defines the two sides: side $\A$
holds the $274$ women and side $\B$ the $277$ men. The protected attribute
is the self-reported race of side $\A$, binarized as the majority category
versus all others. True utility is the mean of both directions' ratings
where both exist, and the available rating otherwise. Across instances, a
mean of $94.7\%$ of directional ratings are missing. Missing ratings are
imputed by row- and column-bias correction, and a small jitter breaks
discrete-rating collisions. The ratings-market qualities are data-derived,
so Assumption~\ref{as:ind} is not guaranteed on that market. Each seed
defines one instance. For the two score-priority markets, each instance
draws a fresh subsample of $400$ applicants. For the ratings market, each
instance uses all $274$ women and draws $274$ of the $277$ men. The
subsample, the interaction-proxy noise, and the encoder initialization all
change with the seed.

\textbf{Encoder architectures:} We encode each market's
raw interaction data using three architectures, all with embedding dimension
$d=16$. \textbf{A0} applies a deterministic rank-$d$ SVD to the observed
interaction matrix. It obtains $x_a$ and $x_b$ from the top-$d$ singular
vectors scaled by the square roots of their singular values. Its diagnostics
vary across instances only because of subsampling and interaction noise.
\textbf{A1} is a two-tower network with separate embedding tables for the
two sides. Each table is passed through a two-layer MLP (linear, ReLU,
linear), and the network is trained with pairwise BPR loss against sampled
negative pairs~\cite{rendle2009bpr}. \textbf{A2} is a bipartite GNN based on
LightGCN~\cite{he2020lightgcn}. It propagates the same embedding tables for
two layers over a degree-normalized bipartite graph, averages the layer
outputs, and uses the BPR procedure of A1. Thus, A1 and A2 differ only in
the function that maps the embedding tables to $x_a$ and $x_b$. Both also
vary across instances because of SGD stochasticity and subsampling.

\textbf{Interaction proxy and BPR pair sampling:}\label{sec:datagen}
For the score-priority markets, Law School and ENEM, applicant scores are
standardized, $\mathrm{score}_z[a]=(\mathrm{score}[a]-\bar{\mathrm{score}})/\mathrm{sd}(\mathrm{score})$,
and shifted to a positive range as Remark~\ref{rem:vertjust} requires. The
quality ladder of side $\B$ is sorted descending from $0.5+1.5\,\epsilon_b$
with $\epsilon_b\sim\mathrm{Uniform}(0,1)$ i.i.d. True utility is the
noiseless outer product $U[a,b]=\mathrm{score}_z[a]\cdot\mathrm{quality}[b]$.
The interaction proxy observed by every encoder, A0, A1, and A2 alike, is
\begin{equation}\label{eq:interaction-proxy}
\mathbf{R}[a,b]=U[a,b]+0.15\,\xi_{a,b}, \qquad \xi_{a,b}\sim\mathcal{N}(0,1)\text{ i.i.d.}
\end{equation}
For the ratings market, $\mathbf{R}$ is the imputed rating matrix defined
above.

A1 and A2 use the same BPR pair-sampling procedure on
$\mathbf{R}$~\cite{rendle2009bpr}. Before training, we compute the fixed
threshold $\mathrm{median}(\mathbf{R})$ and define the positive set
$P=\{(a,b) : \mathbf{R}[a,b]>\mathrm{median}(\mathbf{R})\}$. During each of
$200$ training epochs, we pair every positive example $(a,b^+)\in P$ with
one negative $b^-$ sampled uniformly from the complement of that agent's
positive set. We resample these negatives in every epoch:
\begin{equation}\label{eq:bpr-loss}
\mathcal{L} = -\frac{1}{|P|}\sum_{(a,b^+)\in P} \log\sigma\bigl(s(a,b^+)-s(a,b^-)\bigr),
\end{equation}
optimized by Adam~\cite{kingma2015adam} with learning rate $10^{-2}$. Here
$s(a,b)$ is the score function of Section~\ref{sec:system} with $\phi$ and
$\psi$ taken as the identity, since $x_a,x_b$ are already the final
post-tower representations.

\textbf{Statistical procedures:} All confidence intervals
use a $95\%$ level. The independent market or seed is the sampling unit;
configurations from the same market are not treated as independent, and
encoders fitted to the same market instance form one cluster.
Bootstrap intervals are bias-corrected and accelerated (BCa) using $2000$
resamples of this unit. We report the upper endpoints of two-sided exact
Clopper--Pearson intervals for violation rates and two-sided Wilson score
intervals for false-PASS rates. Paired comparisons use the two-sided
Wilcoxon signed-rank test with the exact null distribution over $10$ seeds.
The associated matched-pairs rank-biserial correlation ranges from $-1$ to
$1$; $\pm1$ indicates that every seed has a difference with the same sign.
The synthetic intervals concern independent market draws from the stated
generator. Realistic-data intervals concern independently randomized runs
conditional on each fixed source dataset.

Every evaluation of Theorem~\ref{thm:main} uses the outlier fraction
$\varepsilon=0.05$, and every evaluation of Theorem~\ref{thm:prob} uses the
failure probability $\eta=0.05$. For a configuration $i$ with realized welfare
change $L_i$ and bound $B_i$, tightness is $L_i/B_i$. Reported mean
tightness is the mean of these configuration-level ratios over
configurations with $B_i>0$. Configurations with $L_i=B_i=0$ are reported
separately and excluded from the tightness average. Let $\mathfrak{M}_n$
denote the set of complete one-to-one matchings. We define the full
feasible welfare range of an instance as
\begin{equation}\label{eq:feasible-range}
\mathcal{R}_{\Wel}
=\max_{\mu\in\mathfrak{M}_n}\Wel(\mu)
-\min_{\mu\in\mathfrak{M}_n}\Wel(\mu).
\end{equation}
Both extrema are computed exactly by solving the maximum-weight and
minimum-weight assignment problems on $U$. A bound is informative when
$B_i<\mathcal{R}_{\Wel}$ for that instance. 

 \textbf{Estimation coordinates:} On the exactly vertical markets, the
constants are computed in the native coordinates $(\theta^\ast,q)$. On the
nominally vertical synthetic markets and the realistic markets, they are
estimated in fixed merit and quality coordinates. The merit coordinate of applicant $a$
is $t_a=n^{-1}\sum_b U(a,b)$. The quality coordinate $\widetilde q$ is the
leading right singular vector of $U-\bar U\mathbf{1}\mathbf{1}^\top$, where
$\bar U$ is the mean entry of $U$, with unit norm and with the sign that
makes its inner product with the column means of $U$ nonnegative.
$L_{\mathrm{mix}}$, $\kappa$, and $\nu$ use $t$; $L_{\mathrm{mix}}$,
$\Lambda$, and $M$ use $\widetilde q$. When $U=s\,q^\top$ for a positive
merit vector $s$, as in the score-priority markets,
the individual constants depend on this choice of coordinates. The
products $M\Lambda$, $M\bar g$, $L_{\mathrm{mix}}\Lambda\kappa$, and
$L_{\mathrm{mix}}\Lambda\nu$ do not. The right-hand sides of
Theorems~\ref{thm:main} and~\ref{thm:prob} and
Proposition~\ref{prop:explicit} therefore coincide with those computed in
the native coordinates $(s,q)$.

\subsection{Screening accuracy on synthetic markets}
\label{sec:cert}

All thresholds are selected on a calibration grid and are held fixed
during evaluation. The grid crosses five merit-correlation levels, five
bias levels, and five seeds, giving $125$ markets that are independent of
the evaluation markets. The gap target is $\tau_{\mathrm{gap}}=0.40$. The
surplus-retention threshold is selected against a reference criterion: a
calibration market receives a reference PASS label when some tested
$\gamma$ closes at least half of the gap while retaining at least $75\%$ of
surplus, and a reference KILL label otherwise. The criterion labels $118$
calibration markets PASS and $7$ KILL, and $\tau_{\mathrm{surp}}=0.90$
attains the highest balanced accuracy against these labels, $0.840$. For the discordance threshold, the
calibration records, for each candidate value, the fraction of nominally
vertical grid markets admitted and the fraction of non-vertical controls
excluded. The threshold is the lowest candidate value that attains the best
balanced accuracy. Discordance runs from $0.177$ to $0.501$ on the grid
markets and equals $1.000$ on every control, so $\tau_{\mathrm{disc}}=0.55$
attains balanced accuracy $1.000$. No grid market attains the exact tier.
The same grid yields a reference entanglement level $\tau_{\ent}=0.50$. It
is used only to interpret verdicts against Proposition~\ref{prop:explicit}
and does not enter the verdict rule.

With these thresholds, the screening verdict agreed with the reference
verdict for all $30$ held-out test markets, with $10$ seeds for each of the
LM-HB, HM-LB, and NV regimes. Table~\ref{tab:cert-confusion} gives the confusion
matrix. The observed accuracy was $1.00$, and the observed false-PASS rate
among the $20$ markets whose reference verdict is not PASS was $0/20$, with a two-sided Wilson $95\%$
upper confidence bound of $0.161$. No NV market received a PASS verdict.

\begin{table}[!t]
\caption{Screening-rule verdicts against the reference verdict of each synthetic regime, with $n=200$ agents per side and 10 held-out seeds per regime (30 markets).}
\label{tab:cert-confusion}
\centering
\footnotesize
\setlength{\tabcolsep}{5pt}
\renewcommand{\arraystretch}{1.05}

\begin{tabular}{lccc}
\toprule
\multirow{2}{*}{Regime (reference)} &
\multicolumn{3}{c}{Screening-rule verdict} \\
\cmidrule(lr){2-4}
 & PASS & KILL & OUT \\
\midrule
LM-HB (PASS) & 10 & 0  & 0 \\
HM-LB (KILL) & 0  & 10 & 0 \\
NV (OUT) & 0  & 0  & 10 \\
\bottomrule
\end{tabular}
\end{table}

As a sensitivity check, the confusion matrix in
Table~\ref{tab:cert-confusion} is unchanged at each of sixteen candidate
values of $\tau_{\mathrm{disc}}$.

\subsection{Evaluation within the exact scope}
\label{sec:exact}

Each regime contains $30$ markets at $n=200$ with ten non-zero edit
strengths, $\gamma\in\{0.1,0.2,\ldots,1.0\}$. On every market and at every
tested $\gamma$, $\text{P2}_\theta=0$ before and after the edit, the orders
of $\theta$, $\theta'$, and $q$ are strict, the stable matchings before and
after the edit are assortative, $U$ is positive and of rank one, and
$\delta_\gamma\ge\max_a|\zeta_a|$. The qualities and the side-$\B$ scale
factors are drawn independently. The pooled whitening map and the crossing
multiplicities therefore do not depend on quality, and
Assumption~\ref{as:ind} holds by the sufficient condition proved in the
supplementary material. Table~\ref{tab:exact} compares the realized
welfare change with the three right-hand sides.

\begin{table}[h!]
\caption{Welfare change against the three bounds on exactly vertical
markets ($n=200$, 30 markets per regime, 10 values of $\gamma$).
Informative is the share of
configurations whose right-hand side is below the full feasible welfare
range in Equation~\eqref{eq:feasible-range}. Violations are counted per
market.}
\label{tab:exact}
\centering
\resizebox{\columnwidth}{!}{%
\renewcommand{\arraystretch}{1.05}
\begin{tabular}{llcccc}
\toprule
Regime & Bound & Mean RHS & Tightness & Informative & Violations \\
\midrule
LM-HB & Theorem~\ref{thm:main} & $2.6\times10^{5}$ & $2.6\times10^{-6}$ & $0.00$ & 0/30 \\
LM-HB & Theorem~\ref{thm:prob} & $0.87$ & $1.0\times10^{-1}$ & $0.69$ & 0/30 \\
LM-HB & Proposition~\ref{prop:explicit} & $1.4\times10^{5}$ & $4.4\times10^{-6}$ & $0.00$ & 0/30 \\
\midrule
HM-LB & Theorem~\ref{thm:main} & $2.6\times10^{5}$ & $3.6\times10^{-7}$ & $0.00$ & 0/30 \\
HM-LB & Theorem~\ref{thm:prob} & $0.38$ & $2.0\times10^{-2}$ & $0.87$ & 0/30 \\
HM-LB & Proposition~\ref{prop:explicit} & $3.1\times10^{5}$ & $3.1\times10^{-7}$ & $0.00$ & 0/30 \\
\bottomrule
\end{tabular}%
}
\end{table}

No bound was violated on any of the $60$ markets, and the exact $95\%$
upper bound on the per-market violation rate is $0.116$ in each regime.
The mean realized welfare change is $0.066$ in the LM-HB regime and
$0.014$ in the HM-LB regime. Within the exact scope the probabilistic bound
is below the feasible welfare range in $68.7\%$ of LM-HB and $87.3\%$ of
HM-LB
configurations. The deterministic bound and the explicit residual bound are
above the feasible welfare range in every configuration.

The smallest margin of $\delta_\gamma$ over $\max_a|\zeta_a|$ is $0.007$.
In the
$18$ configurations where no applicant changes rank, the realized welfare
change and the right-hand side of Theorem~\ref{thm:prob} are both zero,
while the right-hand sides of Theorem~\ref{thm:main} and
Proposition~\ref{prop:explicit} remain positive. Over the remaining
configurations, the right-hand side of Theorem~\ref{thm:prob} exceeds the
realized change by at least $77\%$ of its value.

\subsection{Stress tests outside the exact scope}
\label{sec:bound}

Each regime contains $330$ configurations from $30$ seeds
and $11$ values of $\gamma$, with $n=200$ agents per side. The independent
unit is the market, which is counted as a violation if any tested $\gamma$
violates the expression. The nominally vertical markets do not reach the
exact tier. Their results therefore characterize the plug-in bound
expressions outside the full assumptions of the theorems. The non-vertical
regime provides a further stress test outside
Assumption~\ref{as:vertical}. Neither plug-in expression is violated in any
of the $30$ markets in each regime. The exact $95\%$ upper bound on the
per-market violation rate is $0.116$. Table~\ref{tab:bounds} reports the
mean right-hand side and mean tightness ratio for both expressions.

\begin{table}[!t]
\caption{Mean right-hand side and mean tightness ratio of the worst-case and
probabilistic bounds in the three synthetic regimes ($n=200$ agents per
side, 30 seeds, 11 values of $\gamma$). Worst-case tightness uses all 330
configurations. Probabilistic tightness uses the 300 configurations with
$\gamma>0$, because the probabilistic right-hand side is zero at
$\gamma=0$. NV denotes the non-vertical regime.}
\label{tab:bounds}
\centering
\footnotesize
\setlength{\tabcolsep}{4pt}
\renewcommand{\arraystretch}{1.05}

\begin{tabular}{lcccc}
\toprule
& \multicolumn{2}{c}{Worst-Case} &
\multicolumn{2}{c}{Probabilistic} \\
\cmidrule(lr){2-3}\cmidrule(lr){4-5}
Regime & Mean RHS & Tightness & Mean RHS & Tightness \\
\midrule
LM-HB & $1.10\times10^{5}$ & $2.7\times10^{-6}$ & 1.79 & $7.10\times10^{-2}$ \\
HM-LB & $2.87\times10^{6}$ & $2.6\times10^{-6}$ & 1.86 & $9.36\times10^{-2}$ \\
NV & $7.82\times10^{10}$ & $3.6\times10^{-11}$ & $7.74\times10^{5}$ & $1.7\times10^{-6}$ \\
\bottomrule
\end{tabular}
\end{table}

The worst-case right-hand side exceeds the full feasible
welfare range in every configuration across all three regimes. The
probabilistic refinement improves mean tightness by four to five orders of
magnitude. Under the criterion of Equation~\eqref{eq:feasible-range}, the
probabilistic bound is non-informative in $61.5\%$ of LM-HB
configurations, $53.9\%$ of HM-LB configurations, and $90.9\%$ of NV
configurations. It therefore provides numerical information for $38.5\%$
of the LM-HB and $46.1\%$ of the HM-LB configurations, compared with
$68.7\%$ and $87.3\%$ within the exact scope (Section~\ref{sec:exact}). Mean right-hand side
and tightness separate the NV regime from the two nominally vertical regimes
but not LM-HB from HM-LB. The probabilistic mean right-hand side is $1.79$
for LM-HB and $1.86$ for HM-LB, compared with $7.74\times10^{5}$ for NV, and the corresponding tightness ratios
are $7.1\times10^{-2}$, $9.4\times10^{-2}$, and $1.7\times10^{-6}$.

The expression of Proposition~\ref{prop:explicit} was evaluated on the
same configurations. No violation was observed, and the shift scale of
Equation~\eqref{eq:deltagamma} dominated the realized appeal shift at every
$\gamma$ in every instance. Its mean right-hand side is $6.39\times10^{4}$
for LM-HB, $2.18\times10^{6}$ for HM-LB, and $1.36\times10^{11}$ for NV, which places it beside the worst-case bound and four to six
orders of magnitude above the probabilistic one ($4.6$, $6.1$, and $5.2$
orders in the LM-HB, HM-LB, and NV regimes). The probabilistic expression has
the smallest mean right-hand side in each evaluated regime.

Table~\ref{tab:p2-constants} reports the constants of
Proposition~\ref{prop:explicit}. Two quantities separate the LM-HB and HM-LB
regimes. The merit-appeal coupling $\kappa$ is ten times larger in HM-LB
than in LM-HB, and entanglement is larger in HM-LB as well. The exchange-cost constant
$L_{\mathrm{mix}}$ rises by six orders of magnitude in the non-vertical
control, while the geometric quantities $c$, $A_\parallel$, $B_\perp$, and
$\delta_\gamma$ remain of the same order as in the vertical regimes.

\begin{table}[!t]
\caption{Constants of Proposition~\ref{prop:explicit} in the three
synthetic regimes, as means over 30 seeds. $L_{\mathrm{mix}}$ is the
exchange-cost constant of Assumption~\ref{as:mix}, $\kappa$ and $\nu$ are
the merit-regularity constants of Assumption~\ref{as:merit}, $c$ is the
appeal alignment of the erased direction, and $\delta_\gamma$ is the shift
scale of Equation~\eqref{eq:deltagamma} averaged over the sweep. All are
plug-in estimates, with merit and quality measured in the coordinates
stated in Section~\ref{sec:setup}. NV denotes the non-vertical regime.}
\label{tab:p2-constants}
\centering
\footnotesize
\setlength{\tabcolsep}{4pt}
\renewcommand{\arraystretch}{1.05}
\begin{tabular}{lcccccc}
\toprule
Regime & $L_{\mathrm{mix}}$ & $\kappa$ & $\nu$ & $c$ & $\ent$ & $\delta_\gamma$ \\
\midrule
LM-HB & 4.90 & 0.21 & 6.11 & 1.35 & 0.02 & 4.71 \\
HM-LB & 5.95 & 2.12 & 1.59 & 0.69 & 0.63 & 2.92 \\
NV & $4.7\times10^{7}$ & 0.51 & 0.46 & 0.17 & 0.17 & 0.62 \\
\bottomrule
\end{tabular}
\end{table}

\subsection{Entanglement, assortativity, and alignment}
\label{sec:ent}

Proposition~\ref{prop:explicit} identifies
$\delta_\gamma$ as the shift scale that governs the effective tightness of
the bound. This scale combines entanglement, appeal alignment, and
representation spread. Three analyses isolate these components: the
association between entanglement and welfare loss, the agreement between
assortative prediction and deferred acceptance, and the welfare effect of
aligned versus unaligned perturbations of equal magnitude.

At full erasure, we define welfare loss as
$\max(0,\Wel(\mu)-\Wel(\mu_1))$. Entanglement is positively associated with
this loss, with Spearman $\rho=0.779$ ($p=4.5\times10^{-51}$, $n=245$
across a $7\times7$ grid over $(\omega,\beta)$;
Table~\ref{tab:ent-loss}). Among the $102$ markets with positive loss, the
log-log slope is $0.79$, with a $95\%$ confidence interval of
$[0.64,1.07]$. The slope is taken with respect to entanglement rather than
the shift scale $\delta_\gamma$ of Proposition~\ref{prop:explicit}, so it
does not identify the order of the bound in $\delta_\gamma$.

\begin{table}[!t]
\caption{Relationship between entanglement and welfare loss over the $7\times7$ $(\omega,\beta)$ grid.}
\label{tab:ent-loss}
\centering
\footnotesize
\setlength{\tabcolsep}{5pt}
\renewcommand{\arraystretch}{1.05}

\begin{tabular}{lr}
\toprule
Statistic & Value \\
\midrule
Spearman $\rho$ & 0.779 \\
$p$-value & $4.5\times10^{-51}$ \\
Grid samples ($n$) & 245 \\
Positive-loss samples & 102 \\
Log--log slope & 0.79 \\
95\% CI (slope) & $[0.64,\;1.07]$ \\
\bottomrule
\end{tabular}
\end{table}

\label{sec:assort}
Across a mix of nominally vertical and non-vertical regimes, the assortative
prediction and exact deferred acceptance assigned the same partner in
$26.8\%$ of cases (Table~\ref{tab:assort}). Kendall's $\tau$ on
partner-quality rank was $0.76$. Kendall's $\tau$ tracked the discordance
P2 of Definition~\ref{def:cert}, at Spearman $\rho=0.92$ between $\tau$ and
one minus discordance ($p=3.2\times10^{-17}$), whereas the correlation of
$\tau$ with the rank-one share of Remark~\ref{rem:vertjust} was $0.46$ ($p=0.003$). The assortative approximation
degrades as discordance rises, and discordance predicts that degradation
more closely than the rank-one share does.

\begin{table}[h]
\centering
\caption{Assortative-prediction agreement with exact deferred acceptance,
across mixed nominally vertical and non-vertical regimes.}
\footnotesize
\begin{tabular}{@{}lr@{}}
\toprule
Statistic & Value \\
\midrule
Exact partner-match rate & $26.8\%$ CI $[21.4,31.4]\%$ \\
Kendall's $\tau$ (partner-quality rank) & $0.76$ CI $[0.62,0.87]$ \\
$\tau$ vs.\ $1-$discordance, Spearman $\rho$ & $0.92$ ($p=3.2\times10^{-17}$) \\
$\tau$ vs.\ rank-one share, Spearman $\rho$ & $0.46$ ($p=0.003$) \\
\bottomrule
\end{tabular}

\label{tab:assort}
\end{table}

\label{sec:ablations}
The structural ablation compares coherent rank-one erasure with incoherent
random perturbation of matched magnitude, and one-sided with two-sided
erasure. The two-sided variant, Algorithm~\ref{alg:erase-two}, applies the
same estimated direction to both market sides and is not covered by
Theorem~\ref{thm:main} or Theorem~\ref{thm:prob}.

\begin{algorithm}[ht]
\caption{Two-sided variant used in the structural ablation.}
\label{alg:erase-two}
\begin{algorithmic}[1]
\State run Algorithm~\ref{alg:erase} lines 1--11 to obtain $W$, $W^{-1}$ and $\uhat$
\For{each $a\in\A$ and each $b\in\B$}
  \State $x_a^{w\prime} \gets x_a^w - \gamma\,(x_a^w\cdot\uhat)\,\uhat$;\quad
  $x_b^{w\prime} \gets x_b^w - \gamma\,(x_b^w\cdot\uhat)\,\uhat$
\EndFor
\State \Return $\{W^{-1}x_a^{w\prime}\}$, $\{W^{-1}x_b^{w\prime}\}$
\end{algorithmic}
\end{algorithm}

For perturbations of equal magnitude and side coverage,
coherent rank-one erasure causes more welfare disruption than incoherent
random perturbation. The random-to-structured cost ratio is $0.19$, with
$p=0.002$ and rank-biserial correlation $-1.0$
(Table~\ref{tab:ablations}). This result indicates that alignment with merit,
rather than edit magnitude alone, contributes to cost. One-sided erasure
edits side $\A$ only, as in Algorithm~\ref{alg:erase}, or side $\B$ only;
two-sided erasure is Algorithm~\ref{alg:erase-two}. At $\gamma=0.5$,
two-sided erasure leaves a smaller residual gap than either one-sided
variant, and its gap curve has a smaller area over the full $\gamma$ sweep
($p=0.002$, effect size $1.0$ for both tests).

\begin{table}[h!]
\centering
\caption{Structural ablations: coherent vs.\ incoherent erasure, and one- vs.\ two-sided erasure at $\gamma=0.5$ ($n=10$ seeds, Wilcoxon signed-rank, exact, paired). One-sided $A$ and one-sided $B$ edit only side $\A$ or only side $\B$; two-sided erasure is Algorithm~\ref{alg:erase-two}.}
\label{tab:ablations}
\footnotesize
\begin{tabularx}{\linewidth}{@{}
  >{\hsize=1.15\hsize\raggedright\arraybackslash}X
  >{\hsize=0.85\hsize\raggedright\arraybackslash}X
  >{\hsize=1.0\hsize\raggedright\arraybackslash}X @{}}
\toprule
Comparison & Value & Test statistic \\
\midrule
Random / structured cost ratio & $0.19$ \newline CI $[0.17,0.21]$ & $p=0.002$, \newline rank-biserial $-1.0$ \\
\midrule
Residual gap: one-sided $A$ & $0.116$ & \\
Residual gap: one-sided $B$ & $0.104$ & $p=0.002$, effect $1.0$ ($A$ vs.\ both) \\
Residual gap: two-sided     & $0.070$ & \\
\midrule
Gap AUC: one-sided $A$ & $0.098$ & \\
Gap AUC: one-sided $B$ & $0.086$ & $p=0.002$, effect $1.0$ \\
Gap AUC: two-sided     & $0.064$ & \\
\bottomrule
\end{tabularx}
\end{table}

\subsection{Attribute leakage, retargeting, and computational cost}
\label{sec:nonlinear}
At the PASS operating point, two probes predict the protected attribute
from the edited side-$\A$ representations at the $\gamma$ selected by the
screen: a logistic regression and a multilayer perceptron with one hidden
layer of $32$ units, each scored by three-fold cross-validated AUROC. Across
$15$ seeds, the linear probe attains an AUROC of $1.000$ and the nonlinear
probe attains $0.982$. The nonlinear minus linear difference is $-0.018$, with a
confidence interval of $[-0.048,-0.002]$. These probe results show that a
PASS verdict concerns the empirical gap-closure and surplus-retention
thresholds, not protected-attribute unpredictability.

\label{sec:cost}
Retargeting under partial labels degraded
the estimated direction monotonically, with cosine distance $0$ at full
labels, $0.030$ at $50\%$, $0.203$ at $10\%$, and $0.553$ at $1\%$.

The full edit of Algorithm~\ref{alg:erase}, including whitening, direction
estimation, erasure, and the map back, took $0.79$ ms for $n{=}200$,
$1.25$ ms for $n{=}500$, and $1.82$ ms for $n{=}1000$, against $19$ ms,
$128$ ms, and $531$ ms for one call of deferred acceptance, as medians of
five runs. The projector was therefore not the bottleneck at the evaluated
scales. A rank-$\ell$ variant repeats the direction-estimation step of
Algorithm~\ref{alg:erase} against the residual after removing each
previously estimated direction. Replacing the rank-one edit with rank-$2$
and rank-$4$ variants
left the verdict unchanged on all $10$ LM-HB and all $10$ HM-LB markets
at $n=200$.

\subsection{Realistic datasets}
\label{sec:realdata}

Each of the three markets is encoded with A0, A1, and A2, yielding $30$
instances $\times$ $3$ architectures $\times$ $3$ datasets $=270$
instances. Each instance is tested at ten nonzero values of $\gamma$.
Neither the probabilistic nor the worst-case expression is violated in any of the $30$ runs per dataset, where a run comprises one subsample and the
three encoders fitted to it. The exact $95\%$ upper bound on the per-run
violation probability is $0.116$, conditional on each fixed source dataset.
No real-data cell reaches the
exact tier, so these results characterize the plug-in expressions outside
the full assumptions of the theorems rather than verify the guarantees in
their exact scope.

\begin{figure}[h!]
\centering
\includegraphics[width=0.92\linewidth]{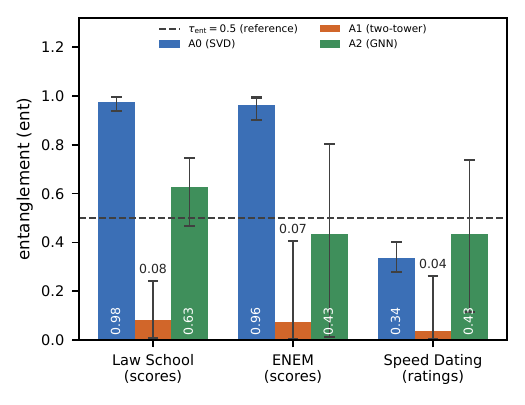}
\caption{Encoder-dependence of the screening rule's entanglement diagnostic, over $30$ instances per cell. Bar heights are means, and whiskers give the observed minimum and maximum across instances. The reference line is $\tau_{\ent}=0.50$. A0's entanglement on the score-priority datasets is near $1$, while the trained encoders spread across nearly the full range on identical data.}
\label{fig:ent-vert}
\end{figure}

\begin{figure}[h!]
\centering
\includegraphics[width=0.92\linewidth]{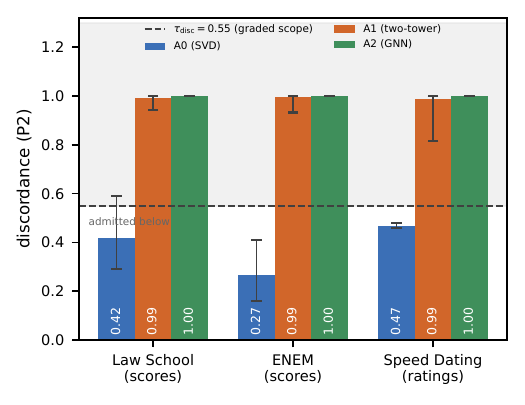}
\caption{Encoder-dependence of the discordance statistic P2, over $30$
instances per cell. Bar heights are means, and whiskers give the observed
minimum and maximum across instances. Lower values indicate closer agreement among side-$\B$ rankings. The
dashed line is the calibrated screen threshold $\tau_{\mathrm{disc}}=0.55$,
and the shaded band above it is the excluded region. No cell attains the exact tier,
which requires a discordance of zero.}
\label{fig:discordance}
\end{figure}

\begin{figure}[h!]
\centering
\includegraphics[width=0.92\linewidth]{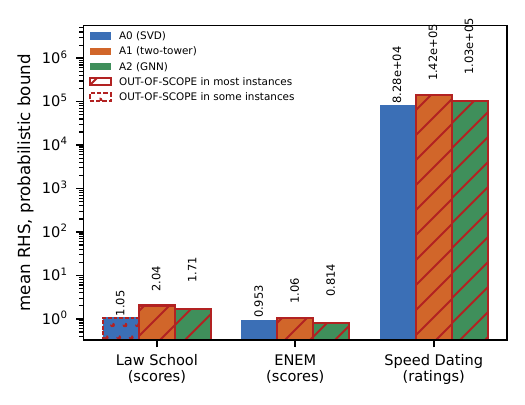}
\caption{Encoder-dependence of the mean right-hand side of the probabilistic bound, log scale, over 30 instances per cell. Bar heights are means over instances. Solid red hatching marks cells that are out of scope in most instances, and dotted red hatching marks cells that are out of scope in some instances.}
\label{fig:bound-magnitude}
\end{figure}

Table~\ref{tab:realdata} and
Figures~\ref{fig:ent-vert}--\ref{fig:bound-magnitude} show that the scope
verdict depends on the encoder even when the data are fixed. For the
deterministic factorization A0, mean discordance ranges from $0.27$ to
$0.47$, and $90\%$ to $100\%$ of instances remain within graded scope in
all three markets. For both trained encoders, mean discordance ranges from
$0.99$ to $1.00$, and every instance is out of scope. Thus, under the two
trained encoders, side-$\B$ rankings do not exhibit the agreement required
by Assumption~\ref{as:vertical}, for either applicant records or observed
ratings. Discordance remains above zero even under A0, so every admitted
real-data cell satisfies only the graded tier.

Discordance and the rank-one share of Remark~\ref{rem:vertjust} disagree in
both directions. Under A1 both score markets have a leading-singular-value share
of $0.99$ with discordance $0.99$: the score matrix is almost exactly rank
one, yet essentially no pair of applicants is ranked unanimously, which is
the sign configuration of Remark~\ref{rem:vertjust}. The ratings market
under A0 is the opposite case, with a share of $0.10$ and discordance
$0.47$, so a rank-one reading would exclude the market that the screen
admits most clearly.

Within a cell, discordance under A0 ranges from $0.29$ to $0.59$ on the
law-school market across instances that differ only in subsample and
interaction noise, which moves that cell out of scope in $10\%$ of
instances, while ENEM A0 stays in scope throughout. Under A1 and A2, SGD
variation moves entanglement across most of its range, as
Figure~\ref{fig:ent-vert} shows, without moving the scope verdict.

Among in-scope cells, PASS rates are not monotonic in
entanglement. A0 passes every ratings instance at $\ent=0.34$ and $70\%$ of
ENEM instances at $\ent=0.96$. In contrast, no law-school instance passes,
although $90\%$ of that cell is in scope. Passing instances use partial
edits, with mean selected $\gamma$ values of $0.56$ for ENEM and $0.73$ for
the ratings market. Section~\ref{sec:disc-ent} interprets this pattern using
Proposition~\ref{prop:explicit}.

On the real markets the merit-appeal coupling $\kappa$ tracks entanglement
across the nine cells, from $2.00$ at $\ent=0.96$ down to $0.06$ at
$\ent=0.08$, and no violation of the expression of
Proposition~\ref{prop:explicit} was observed in any of the $2700$
configurations. Its right-hand side lies between
$1.1\times10^{5}$ and $2.1\times10^{11}$, driven by $L_{\mathrm{mix}}$,
which is $7.33$ on the law-school market, $7.37$ on the national-exam
market, and $8.0\times10^{5}$ on the ratings market, as means over
instances.  These values use the estimation coordinates of Section~\ref{sec:setup}.

Figure~\ref{fig:bound-magnitude} plots the mean probabilistic RHS and marks
the cells whose majority verdict is OUT-OF-SCOPE. Magnitude does not
identify those cells. On the ratings market the in-scope A0 cell gives a
mean RHS of $8.3\times10^{4}$ and the out-of-scope A1 cell gives
$1.4\times10^{5}$. On the score-priority markets the bound spans $0.81$ to
$2.04$ across cells that are variously in and out of scope, five orders of
magnitude below the ratings cells. Magnitude tracks the market rather than
the scope verdict.

\begin{table*}[!t]
\caption{Summary over three realistic markets and three architectures, 30
instances per cell. Each instance draws its own subsample, interaction
noise, and encoder initialization. Ent.\ and Mean RHS are means over
instances. P2 is the discordance of Definition~\ref{def:cert}, with its observed
minimum and maximum across instances, and the next column is the leading-singular-value share of
Remark~\ref{rem:vertjust}. Lower discordance indicates closer agreement
among side-$\B$ rankings, and the graded
screen admits an instance at $\text{P2}\le0.55$. PASS and OUT are the fractions
of instances receiving those verdicts; the remainder are KILL. Mean RHS is
the probabilistic right-hand side, and the next column is the right-hand
side of Proposition~\ref{prop:explicit} on the same instances. Viol.\
counts violations of the probabilistic bound over the ten non-zero values
of $\gamma$ in each instance.}
\label{tab:realdata}
\centering
\footnotesize
\setlength{\tabcolsep}{7pt}
\renewcommand{\arraystretch}{1.08}

\begin{tabular}{llcccccccc}
\toprule
Dataset & Arch. & Ent. & P2, discordance (range) & Rank-1 share & PASS &
OUT & Mean RHS & Prop.~\ref{prop:explicit} RHS & Viol. \\
\midrule
Law School & A0 & 0.98 & 0.42 [0.29, 0.59] & 0.99 & 0.00 & 0.10 & 1.05 & $2.6\times10^{7}$ & 0/300 \\
Law School & A1 & 0.08 & 0.99 [0.94, 1.00] & 0.99 & 0.00 & 1.00 & 2.04 & $2.3\times10^{5}$ & 0/300 \\
Law School & A2 & 0.63 & 1.00 [1.00, 1.00] & 0.87 & 0.00 & 1.00 & 1.71 & $8.0\times10^{5}$ & 0/300 \\
\midrule
ENEM & A0 & 0.96 & 0.27 [0.16, 0.41] & 0.99 & 0.70 & 0.00 & 0.95 & $2.5\times10^{5}$ & 0/300 \\
ENEM & A1 & 0.07 & 1.00 [0.93, 1.00] & 0.99 & 0.00 & 1.00 & 1.06 & $1.1\times10^{5}$ & 0/300 \\
ENEM & A2 & 0.43 & 1.00 [1.00, 1.00] & 0.82 & 0.00 & 1.00 & 0.81 & $1.1\times10^{5}$ & 0/300 \\
\midrule
Ratings & A0 & 0.34 & 0.47 [0.46, 0.48] & 0.10 & 1.00 & 0.00 & $8.3\times10^{4}$ & $2.1\times10^{11}$ & 0/300 \\
Ratings & A1 & 0.04 & 0.99 [0.82, 1.00] & 0.75 & 0.00 & 1.00 & $1.4\times10^{5}$ & $9.9\times10^{10}$ & 0/300 \\
Ratings & A2 & 0.43 & 1.00 [1.00, 1.00] & 0.67 & 0.00 & 1.00 & $1.0\times10^{5}$ & $6.4\times10^{9}$ & 0/300 \\
\bottomrule
\end{tabular}
\end{table*}

\subsection{Comparison to prior post-hoc interventions}
\label{sec:sota}

We compare the proposed method with iterative nullspace
projection~\cite{ravfogel2020null} and three simplified reimplementations of
prior approaches. The RLACE-style baseline alternates a linear adversary
with an exponential-moving-average update of the erased direction rather
than using the constrained minimax solver in~\cite{ravfogel2022linear}.
Group score standardization rescales each group's appeal scores to the
pooled mean and standard deviation, following the general post-processing
principle of~\cite{hardt2016equality}, without outcome labels. The
reject-option-style baseline moves disadvantaged-group scores within a band
around the pooled median halfway toward the advantaged-group mean, following
\cite{kamiran2012decision}. All four baselines operate at inference time on
a fixed encoder and use the same market, mechanism, and metrics.
Tables~\ref{tab:sota_pass} and~\ref{tab:sota_kill} report the fraction of
the group gap closed and the welfare retained relative to the uncorrected
system. For the proposed method, the screening rule selects both $\gamma$
and the verdict. We evaluate the LM-HB and HM-LB regimes of Section~\ref{sec:setup}, each
with $n=400$ agents per side and $30$ seeds. Because the proposed method is
a screening and selection procedure, it applies the smallest tested edit
that meets both targets under a PASS verdict, or the gap target alone under
a KILL verdict. Each baseline is evaluated at its standard full-correction
operating point. The comparison therefore assesses the procedures as
deployed, not interventions of equal strength.

In the LM-HB regime, the screening rule returns PASS for
all $30$ seeds and selects a mean $\gamma$ of $0.89$. At the selected edit,
the proposed method closes $61.1\%$ of the gap and retains $102.4\%$ of
welfare, corresponding to a $2.4\%$ increase relative to the uncorrected
matching. Group score standardization closes $97.0\%$ of the gap, and
iterative nullspace projection closes $69.9\%$, both with comparable
welfare retention. The RLACE-style and reject-option-style baselines close
$36.5\%$ and $5.5\%$, respectively.

\begin{table}[!t]
\caption{Comparison with post-hoc baselines on the LM-HB market ($\omega=0.05$, $\beta=0.65$, $n=400$, 30 seeds), relative to the uncorrected system. Values are means over seeds. Welfare retention above $100\%$ means that welfare improved relative to the uncorrected matching. RLACE-style, group score standardization, and reject-option-style are simplified reimplementations.}
\label{tab:sota_pass}
\centering
\footnotesize
\setlength{\tabcolsep}{3.5pt}
\renewcommand{\arraystretch}{1.05}
\resizebox{\columnwidth}{!}{%
\begin{tabular}{lcccc}
\toprule
Method & Gap (\%) & Welfare (\%) & Verdict & Retrain \\
\midrule
Ours (with verdict) & 61.1 & 102.4 & Yes & No \\
INLP \cite{ravfogel2020null} & 69.9 & 102.5 & No & No \\
RLACE-style \cite{ravfogel2022linear} & 36.5 & 101.7 & No & No \\
Group score standardization \cite{hardt2016equality} & 97.0 & 101.2 & No & No \\
Reject-option-style \cite{kamiran2012decision} & 5.5 & 98.2 & No & No \\
\bottomrule
\end{tabular}%
}
\end{table}

In the HM-LB regime, the protected direction is strongly
entangled with true merit at $\omega=0.65$. The screen returns KILL for $8$
seeds and OUT-OF-SCOPE for $22$. The scope verdict in this regime depends on
market size. With the generator and thresholds fixed, the discordance used
by the screen, the maximum over the tested $\gamma$, has mean $0.32$ at
$n=100$, $0.44$ at $n=200$, $0.55$ at $n=300$, and $0.60$ at $n=400$, and
the share of seeds above $\tau_{\mathrm{disc}}$ is $0.00$, $0.03$, $0.47$,
and $0.73$. Pre-edit discordance stays between $0.11$ and $0.15$ at every
size, so the exclusions arise in the edited markets. The ten HM-LB markets of Table~\ref{tab:cert-confusion} at $n=200$ all lie below the
threshold. In the LM-HB regime, the same quantity stays between $0.22$ and
$0.27$, and every seed receives PASS at every size.

For the $8$ seeds receiving KILL, no tested
$\gamma$ both closes $\tau_{\mathrm{gap}}$ of the gap and retains
$\tau_{\mathrm{surp}}$ of the surplus. At the smallest $\gamma$ that meets
the gap target, whose mean is $0.95$, the method closes $85.6\%$ of the gap
and retains $95.0\%$ of welfare. INLP closes $6.3\%$ of the gap with
$100.1\%$ welfare retention. The RLACE-style baseline closes $47.1\%$ with
$87.8\%$ welfare retention, group score standardization closes $98.0\%$
with $96.9\%$ retention, and the reject-option-style baseline closes
$25.3\%$.

\begin{table}[!t]
\caption{Comparison with post-hoc baselines on the HM-LB market ($\omega=0.65$, $\beta=0.05$, $n=400$, 30 seeds), relative to the uncorrected system. The row for the proposed method averages the $8$ seeds returning KILL; baseline rows average all $30$ seeds. RLACE-style, group score standardization, and reject-option-style are simplified reimplementations.}
\label{tab:sota_kill}
\centering
\footnotesize
\setlength{\tabcolsep}{3.5pt}
\renewcommand{\arraystretch}{1.05}
\resizebox{\columnwidth}{!}{%
\begin{tabular}{lcccc}
\toprule
Method & Gap (\%) & Welfare (\%) & Verdict & Retrain \\
\midrule
Ours (with verdict) & 85.6 & 95.0 & Yes & No \\
INLP \cite{ravfogel2020null} & 6.3 & 100.1 & No & No \\
RLACE-style \cite{ravfogel2022linear} & 47.1 & 87.8 & No & No \\
Group score standardization \cite{hardt2016equality} & 98.0 & 96.9 & No & No \\
Reject-option-style \cite{kamiran2012decision} & 25.3 & 99.0 & No & No \\
\bottomrule
\end{tabular}%
}
\end{table}

On the same $30$ LM-HB seeds, group score standardization closes
more of the gap than the proposed edit at comparable welfare retention.
In the HM-LB regime, the proposed edit is averaged over the eight seeds receiving KILL,
while baseline averages cover all $30$ seeds; those means do not establish
a paired method comparison. The property that distinguishes the proposed
method is the pre-deployment verdict. It
returned PASS on all $30$ LM-HB seeds and on none of the HM-LB seeds. None of the evaluated baselines reports a verdict or a welfare-cost
bound for its edit.

\subsection{Synthesis}
\label{sec:synthesis}

\begin{table*}[h!]
\caption{Principal empirical claims and their quantitative evidence.}
\label{tab:synthesis-pooled}
\centering
\footnotesize
\setlength{\tabcolsep}{5pt}
\renewcommand{\arraystretch}{1.05}

\begin{tabular}{p{0.22\textwidth}p{0.17\textwidth}p{0.53\textwidth}}
\toprule
Claim & Test / Metric & Key Evidence \\
\midrule

Screening accuracy on labeled markets &
Wilson 95\% CI &
0/20 false PASS among markets with a non-PASS reference verdict; 30/30 verdicts agree with the reference regime verdict; upper 95\% bound = 0.161 \\

Scope test detects sign failures &
Discordance vs.\ rank-one share &
Rank-one share 0.99 with discordance 0.99 (Law School and ENEM under A1); share 0.10 with discordance 0.47 (Ratings under A0) \\

Entanglement is associated with welfare loss &
Spearman, bootstrap &
$\rho=0.779$, $p=4.5\times10^{-51}$; log--log slope $0.79$ (95\% CI $[0.64,1.07]$) \\

Residual-bound expression (Prop.~\ref{prop:explicit}) &
Violation count; dominance check &
No observed violation in 990 synthetic configurations across three regimes and 2700 real-data configurations; $\delta_\gamma$ dominated the realized shift in every instance \\

No bound violation within the exact scope &
Violation count &
0/60 exactly vertical markets violate any evaluated bound; probabilistic bound informative in 68.7\% (LM-HB) and 87.3\% (HM-LB) of configurations \\

Assortativity tracks discordance &
Kendall $\tau$, Spearman &
$\tau=0.76$ (95\% CI $[0.62,0.87]$); $\rho=0.92$ with $1-$discordance, $p=3.2\times10^{-17}$ \\

Aligned erasure costs more than random perturbation &
Wilcoxon signed-rank &
Cost ratio $0.19$ (95\% CI $[0.17,0.21]$); $p=0.002$; rank-biserial $=-1.0$ \\

Two-sided erasure closes the gap faster &
Wilcoxon; AUC &
Residual gap $0.070$ vs.\ $0.116/0.104$; $p=0.002$; effect size $=1.0$ \\

A PASS verdict does not imply attribute unpredictability &
AUROC; bootstrap &
Linear AUROC $1.000$; nonlinear AUROC $0.982$; difference $-0.018$ (95\% CI $[-0.048,-0.002]$) \\

Retargeting is label-gated &
Trend analysis &
Cosine distance increased monotonically from 0 to 0.553 as label availability decreased \\

Projector is not the bottleneck &
Wall-clock timing &
Projector: 0.8--1.8 ms; matching: 19--531 ms at the evaluated sizes \\

\bottomrule
\end{tabular}
\end{table*}

The empirical results support three findings. First, the
screen agrees with the reference verdicts of the pre-specified synthetic
regimes.
Second, the probabilistic expression has the smallest mean right-hand side
among the three evaluated bounds in every regime. Third, the realistic-data
scope verdict depends on both the encoder and the market. The trained
encoders yield near-maximal discordance, whereas the closed-form
factorization retains lower discordance and usually remains within graded
scope. Tables~\ref{tab:bounds}, \ref{tab:realdata}, and
\ref{tab:synthesis-pooled} provide the corresponding quantitative evidence.

\section{Discussion}
\label{sec:discussion}

\subsection{Interpretation of the welfare bounds}
\label{sec:disc-bound}

The two welfare guarantees serve distinct roles.
Theorem~\ref{thm:main} is deterministic under its stated assumptions.
Theorem~\ref{thm:prob} replaces the outlier-dominated constant with a
deviation term and a nonzero failure probability under the exchangeability
assumption, which can yield a tighter finite-sample bound.

At the evaluated scales, the plug-in constants place the
deterministic right-hand side above the feasible welfare range. The
probabilistic refinement is numerically informative for $68.7\%$ to
$87.3\%$ of the configurations within the exact scope
(Section~\ref{sec:exact}) and for $38.5\%$ to $46.1\%$ of the nominally
vertical configurations (Section~\ref{sec:bound}).

Bound magnitude is interpretable only jointly with the scope verdict. On
the synthetic markets, mean right-hand side and mean tightness distinguish
the NV regime from the two nominally vertical regimes, even where
the binary informative criterion does not. On the realistic markets,
magnitude tracks the market rather than the scope verdict
(Section~\ref{sec:realdata}). A bound computed outside
Assumption~\ref{as:vertical} carries no guarantee, whether it is large or
small. Scope is therefore reported as a separate field, and magnitude
serves as a continuous severity indicator only within scope.

\subsection{Screen availability as a model-selection criterion}
\label{sec:disc-modelsel}

The three encoders represent common classes of production
systems. A0 represents classical matrix-factorization recommenders and
score-priority allocation systems, such as admissions based on standardized
tests. A1 represents two-tower retrieval architectures that combine user
and item towers at serving time. A2 represents graph-based recommenders that
propagate information over an interaction graph. The closed-form
factorization remains within graded scope for $90\%$ to $100\%$ of
instances across the three markets, while both trained encoders are out of
scope in every instance, despite using the same data, attribute, and target. These results suggest that a fairness audit may not transfer across
architectures. Applying the screen to candidate
representations before architecture selection would indicate whether the
edit is admissible under the chosen empirical thresholds. This result can
then be considered with predictive performance and latency.

\subsection{Entanglement as a diagnostic rather than a gate}
\label{sec:disc-ent}

For the in-scope real-data cells, PASS rates are not
monotonic in entanglement (Section~\ref{sec:realdata}).
Proposition~\ref{prop:explicit} bounds the welfare change at each edit
strength $\gamma$, while Section~\ref{sec:ent} evaluates full erasure.
The screen selects the smallest tested edit meeting both gap-closure and
surplus-retention targets. Passing instances use partial
edits that retain most of the assortative surplus. High entanglement
therefore indicates that full erasure may be costly, but it does not imply
that every gap-closing partial edit is costly.

Proposition~\ref{prop:explicit} also shows that entanglement alone does not
determine the cost of an edit. The measurable shift scale
$\delta_\gamma$ in Equation~\eqref{eq:deltagamma} depends on
$c=|\ip{W^{-1}\uhat}{k}|$ and the perpendicular spread $B_\perp$, in
addition to $\ent$. Unless representation spread is concentrated along the
merit direction, entanglement alone does not determine the bound. The
surplus sweep instead measures the realized cost of each candidate edit,
and the verdict uses this measurement. For deployments that require full
erasure rather than a target-driven partial edit, entanglement remains a
directly relevant diagnostic.

\subsection{Scope and limitations}
\label{sec:disc-limits}

The guarantees are market-design results based on order statistics, ranks,
and a Lipschitz utility function. They are independent of the training loss,
optimization procedure, and encoder construction. Their scope is limited to
inference-time sensitivity: they neither smooth $\mathcal{M}$ nor provide a
surrogate gradient for end-to-end training through the mechanism. The
locality argument in Lemma~\ref{lem:disp} applies to the rank-one
perturbation in Equation~\eqref{eq:eta} and does not directly cover an
unconstrained multidirectional edit such as fine-tuning.

The exact guarantees require a vertical market with a common ranking
criterion before and after editing. Substantially idiosyncratic preferences,
as in residency matching and freelance platforms, violate this structure.
The analysis also assumes a balanced, one-to-one, complete market;
capacities, unmatched agents, and incomplete preference lists are outside
its scope. The graded tier admits near-vertical markets through a calibrated
discordance threshold but has no corresponding theorem. In that tier, the
screening verdict is supported by the empirical surplus-retention sweep,
which re-solves the mechanism exactly.

The intervention removes one linear mean-difference direction and does not
eliminate nonlinear protected-attribute information or information
distributed across several directions. The screening rule also requires
estimates of utility and partner quality to evaluate surplus. Numerical use
of the deterministic guarantee requires valid upper estimates of the
population constants $M$, $\Lambda$, and $\bar f$. The probabilistic
refinement additionally requires exchangeability of quality-gap positions
relative to the displacement pattern, which may fail for data-derived
quality ladders.

Within this scope, Theorems~\ref{thm:main} and~\ref{thm:prob} bound the
welfare effect of editing the output of a trained encoder, and
Proposition~\ref{prop:explicit} identifies the measurable quantities that
control the cost. The screening rule connects these analytical conditions
to the exact output of the allocation mechanism and separates exact-tier
guarantees from graded empirical decisions.

\section{Conclusion and future work}

\label{sec:conclusion}

We studied whether an inference-time edit can attenuate the
influence of a protected attribute on learned two-sided matching without
retraining and with a cost that can be checked before deployment. For
markets that remain vertically ordered before and after the edit, the
welfare change caused by a closed-form projector is bounded by the
edit-induced appeal shift and an explicit finite-sample and outlier
remainder (Theorem~\ref{thm:main}). Under an additional exchangeability
assumption, a probabilistic refinement improves finite-sample tightness by
several orders of magnitude (Theorem~\ref{thm:prob}). These results
underlie a three-outcome screening rule. On $30$ controlled synthetic markets, the
screen agrees with every pre-specified regime verdict, with no observed
false PASS. On exactly vertical synthetic markets, no evaluated bound is
violated, and the probabilistic bound is informative in $69\%$ to $87\%$
of configurations. Across $270$ instances from three realistic markets, the
closed-form factorization remains within graded scope in at least $90\%$
of instances, whereas both trained encoders are out of scope in every
instance. The realistic markets do not reach the exact tier, so their
evaluations are stress tests. On such markets the screen supports
model-selection and retraining audits under the graded criterion.

Future work should first tighten the constants or develop
high-probability surrogates for the supremum estimates in
Proposition~\ref{prop:explicit}. A second direction is to generalize
Theorems~\ref{thm:main} and~\ref{thm:prob} beyond rank-one edits and derive
an analytical guarantee for the rank-$\ell$ variant. A third direction is
to identify weaker structural assumptions that support closed-form bounds
in markets with idiosyncratic preferences, such as residency matching.
Finally, the observed dependence on encoder architecture motivates training
objectives that preserve a common ranking by construction. Such objectives
could make screen compatibility a model-design constraint rather than only
a post-training diagnostic.


\section*{Declaration of competing interest}
The author declares no known competing financial interests or personal
relationships that could have appeared to influence the work reported in
this paper.

\section*{Data availability}
The law-school market is built from the LSAC National Longitudinal Bar
Passage Study~\cite{wightman1998lsac}, which is available from the Law
School Admission Council under its own access conditions. The
national-exam market is built from the public ENEM
microdata~\cite{inep2023enem}. The ratings market is built from the public
Speed Dating experiment data~\cite{fisman2006gender}. No new data were
collected for this study.  Every market
used here can be reproduced from the cited sources and the
seed values recorded in the configuration files. These files will be made
publicly available upon publication.
\section*{Code availability}
The implementation, the configuration files, and the raw result files
supporting all reported values will be made publicly available in an
archival repository upon publication.

\section*{Declaration of generative AI and AI-assisted technologies in the writing process}
During the preparation of this work, the author used Grammarly and Paperpal to improve language clarity and
readability. After using these tools, the
authors reviewed and edited the content as needed and takes full
responsibility for the content of the published article.

\bibliographystyle{elsarticle-num}
\bibliography{ref}

@INPROCEEDINGS {9407135,
author = { Acun, Bilge and Murphy, Matthew and Wang, Xiaodong and Nie, Jade and Wu, Carole-Jean and Hazelwood, Kim },
booktitle = { 2021 IEEE International Symposium on High-Performance Computer Architecture (HPCA) },
title = {{ Understanding Training Efficiency of Deep Learning Recommendation Models at Scale }},
year = {2021},
volume = {},
ISSN = {},
pages = {802-814},
doi = {10.1109/HPCA51647.2021.00072},
url = {https://doi.ieeecomputersociety.org/10.1109/HPCA51647.2021.00072},
publisher = {IEEE Computer Society},
address = {Los Alamitos, CA, USA},
month =mar}

@inproceedings{10.1145/3219819.3219890,
author = {Ying, Rex and He, Ruining and Chen, Kaifeng and Eksombatchai, Pong and Hamilton, William L. and Leskovec, Jure},
title = {Graph Convolutional Neural Networks for Web-Scale Recommender Systems},
year = {2018},
isbn = {9781450355520},
publisher = {Association for Computing Machinery},
address = {New York, NY, USA},
url = {https://doi.org/10.1145/3219819.3219890},
doi = {10.1145/3219819.3219890},
booktitle = {Proceedings of the 24th ACM SIGKDD International Conference on Knowledge Discovery \& Data Mining},
pages = {974–983},
numpages = {10},
location = {London, United Kingdom},
series = {KDD '18}
}

@article{fisman2006gender,
  author  = {Fisman, Raymond and Iyengar, Sheena S. and Kamenica, Emir and Simonson, Itamar},
  title   = {Gender Differences in Mate Selection: Evidence from a Speed Dating Experiment},
  journal = {The Quarterly Journal of Economics},
  volume  = {121},
  number  = {2},
  pages   = {673--697},
  year    = {2006},
  doi     = {10.1162/qjec.2006.121.2.673}
}

@inproceedings{belrose2023leace,
  author    = {Belrose, Nora and Schneider-Joseph, David and Ravfogel, Shauli and Cotterell, Ryan and Raff, Edward and Biderman, Stella},
  title     = {{LEACE}: Perfect Linear Concept Erasure in Closed Form},
  booktitle = {Advances in Neural Information Processing Systems 36 (NeurIPS 2023)},
  year      = {2023},
  address   = {New Orleans, LA, USA},
}

@article{gale1962college,
  author  = {Gale, David and Shapley, Lloyd S.},
  title   = {College Admissions and the Stability of Marriage},
  journal = {American Mathematical Monthly},
  volume  = {69},
  number  = {1},
  pages   = {9--15},
  year    = {1962},
  doi     = {10.2307/2312726},
}

@inproceedings{hardt2016equality,
  author    = {Hardt, Moritz and Price, Eric and Srebro, Nathan},
  title     = {Equality of Opportunity in Supervised Learning},
  booktitle = {Advances in Neural Information Processing Systems 29 (NeurIPS 2016)},
  pages     = {3323--3331},
  year      = {2016},
}

@inproceedings{he2020lightgcn,
  author    = {He, Xiangnan and Deng, Kuan and Wang, Xiang and Li, Yan and Zhang, Yongdong and Wang, Meng},
  title     = {{LightGCN}: Simplifying and Powering Graph Convolution Network for Recommendation},
  booktitle = {Proceedings of the 43rd International ACM SIGIR Conference on Research and Development in Information Retrieval (SIGIR '20)},
  pages     = {639--648},
  year      = {2020},
  doi       = {10.1145/3397271.3401063},
}

@misc{inep2023enem,
  author       = {{Instituto Nacional de Estudos e Pesquisas Educacionais Anísio Teixeira (INEP)}},
  title        = {Microdados do Exame Nacional do Ensino Médio ({ENEM})},
  year         = {2023},
  howpublished = {Ministério da Educação, Brasília},
  url          = {https://www.gov.br/inep/pt-br/acesso-a-informacao/dados-abertos/microdados},
}

@inproceedings{kamiran2012decision,
  author    = {Kamiran, Faisal and Karim, Asim and Zhang, Xiangliang},
  title     = {Decision Theory for Discrimination-Aware Classification},
  booktitle = {2012 IEEE 12th International Conference on Data Mining (ICDM)},
  pages     = {924--929},
  year      = {2012},
  doi       = {10.1109/ICDM.2012.45},
}

@inproceedings{kingma2015adam,
  author    = {Kingma, Diederik P. and Ba, Jimmy},
  title     = {Adam: A Method for Stochastic Optimization},
  booktitle = {3rd International Conference on Learning Representations (ICLR)},
  year      = {2015},
}

@inproceedings{ravfogel2020null,
  author    = {Ravfogel, Shauli and Elazar, Yanai and Gonen, Hila and Twiton, Michael and Goldberg, Yoav},
  title     = {Null It Out: Guarding Protected Attributes by Iterative Nullspace Projection},
  booktitle = {Proceedings of the 58th Annual Meeting of the Association for Computational Linguistics (ACL)},
  pages     = {7237--7256},
  year      = {2020},
  doi       = {10.18653/v1/2020.acl-main.647},
}

@inproceedings{ravfogel2022linear,
  author    = {Ravfogel, Shauli and Twiton, Michael and Goldberg, Yoav and Cotterell, Ryan},
  title     = {Linear Adversarial Concept Erasure},
  booktitle = {Proceedings of the 39th International Conference on Machine Learning (ICML)},
  volume    = {162},
  pages     = {18400--18421},
  year      = {2022},
}

@inproceedings{rendle2009bpr,
  author    = {Rendle, Steffen and Freudenthaler, Christoph and Gantner, Zeno and Schmidt-Thieme, Lars},
  title     = {{BPR}: Bayesian Personalized Ranking from Implicit Feedback},
  booktitle = {Proceedings of the Twenty-Fifth Conference on Uncertainty in Artificial Intelligence (UAI '09)},
  pages     = {452--461},
  year      = {2009},
}

@book{roth1990two,
  author    = {Roth, Alvin E. and Sotomayor, Marilda A. Oliveira},
  title     = {Two-Sided Matching: A Study in Game-Theoretic Modeling and Analysis},
  series    = {Econometric Society Monographs},
  publisher = {Cambridge University Press},
  address   = {Cambridge, UK},
  year      = {1990},
}

@techreport{wightman1998lsac,
  author      = {Wightman, Linda F.},
  title       = {{LSAC} National Longitudinal Bar Passage Study},
  institution = {Law School Admission Council},
  type        = {{LSAC Research Report Series}},
  year        = {1998},
}

@inproceedings{wu2021fairrec,
  author    = {Wu, Chuhan and Wu, Fangzhao and Wang, Xiting and Huang, Yongfeng and Xie, Xing},
  title     = {Fairness-Aware News Recommendation with Decomposed Adversarial Learning},
  booktitle = {Proceedings of the AAAI Conference on Artificial Intelligence},
  volume    = {35},
  number    = {5},
  pages     = {4462--4469},
  year      = {2021},
}

@inproceedings{wu2021fairgo,
  author    = {Wu, Le and Chen, Lei and Shao, Pengyang and Hong, Richang and Wang, Xiting and Wang, Meng},
  title     = {Learning Fair Representations for Recommendation: A Graph-Based Perspective},
  booktitle = {Proceedings of the Web Conference 2021 (WWW '21)},
  pages     = {2198--2208},
  year      = {2021},
  doi       = {10.1145/3442381.3450015},
}

@inproceedings{zehlike2017fair,
  author    = {Zehlike, Meike and Bonchi, Francesco and Castillo, Carlos and Hajian, Sara and Megahed, Mohamed and Baeza-Yates, Ricardo},
  title     = {{FA*IR}: A Fair Top-k Ranking Algorithm},
  booktitle = {Proceedings of the 2017 ACM on Conference on Information and Knowledge Management (CIKM '17)},
  pages     = {1569--1578},
  year      = {2017},
  doi       = {10.1145/3132847.3132938},
}

@article{roth2008deferred,
  title={Deferred acceptance algorithms: history, theory, practice, and open questions},
  author={Roth, Alvin E},
  journal={International Journal of Game Theory},
  volume={36},
  number={3-4},
  pages={537--569},
  year={2008},
  publisher={Springer},
  doi={10.1007/s00182-008-0117-6}
}

@misc{ravindranath2023deeplearningtwosidedmatching,
      title={Deep Learning for Two-Sided Matching}, 
      author={Sai Srivatsa Ravindranath and Zhe Feng and Shira Li and Jonathan Ma and Scott D. Kominers and David C. Parkes},
      year={2023},
      eprint={2107.03427},
      archivePrefix={arXiv},
      primaryClass={cs.GT},
      url={https://arxiv.org/abs/2107.03427}, 
}

@misc{du2021fairnessrepresentationneutralization,
      title={Fairness via Representation Neutralization}, 
      author={Mengnan Du and Subhabrata Mukherjee and Guanchu Wang and Ruixiang Tang and Ahmed Hassan Awadallah and Xia Hu},
      year={2021},
      eprint={2106.12674},
      archivePrefix={arXiv},
      primaryClass={cs.LG},
      url={https://arxiv.org/abs/2106.12674}, 
}

@inproceedings{bolukbasi2016man,
author = {Bolukbasi, Tolga and Chang, Kai-Wei and Zou, James and Saligrama, Venkatesh and Kalai, Adam},
title = {Man is to computer programmer as woman is to homemaker? debiasing word embeddings},
year = {2016},
isbn = {9781510838819},
publisher = {Curran Associates Inc.},
address = {Red Hook, NY, USA},
booktitle = {Proceedings of the 30th International Conference on Neural Information Processing Systems},
pages = {4356–4364},
numpages = {9},
location = {Barcelona, Spain},
series = {NIPS'16}
}

@inproceedings{pleiss2017fairness,
author = {Pleiss, Geoff and Raghavan, Manish and Wu, Felix and Kleinberg, Jon and Weinberger, Kilian Q.},
title = {On fairness and calibration},
year = {2017},
isbn = {9781510860964},
publisher = {Curran Associates Inc.},
address = {Red Hook, NY, USA},
booktitle = {Proceedings of the 31st International Conference on Neural Information Processing Systems},
pages = {5684–5693},
numpages = {10},
location = {Long Beach, California, USA},
series = {NIPS'17}
}

@InProceedings{zemel2013learning,
  title = 	 {Learning Fair Representations},
  author = 	 {Zemel, Rich and Wu, Yu and Swersky, Kevin and Pitassi, Toni and Dwork, Cynthia},
  booktitle = 	 {Proceedings of the 30th International Conference on Machine Learning},
  pages = 	 {325--333},
  year = 	 {2013},
  editor = 	 {Dasgupta, Sanjoy and McAllester, David},
  volume = 	 {28},
  number =       {3},
  series = 	 {Proceedings of Machine Learning Research},
  address = 	 {Atlanta, Georgia, USA},
  month = 	 {17--19 Jun},
  publisher =    {PMLR},
  url = 	 {https://proceedings.mlr.press/v28/zemel13.html}
}

@inproceedings{yao2017beyond,
author = {Yao, Sirui and Huang, Bert},
title = {Beyond parity: fairness objectives for collaborative filtering},
year = {2017},
isbn = {9781510860964},
publisher = {Curran Associates Inc.},
address = {Red Hook, NY, USA},
booktitle = {Proceedings of the 31st International Conference on Neural Information Processing Systems},
pages = {2925–2934},
numpages = {10},
location = {Long Beach, California, USA},
series = {NIPS'17}
}

@article{kamada2015efficient,
Author = {Kamada, Yuichiro and Kojima, Fuhito},
Title = {Efficient Matching under Distributional Constraints: Theory and Applications},
Journal = {American Economic Review},
Volume = {105},
Number = {1},
Year = {2015},
Month = {January},
Pages = {67–99},
DOI = {10.1257/aer.20101552},
URL = {https://www.aeaweb.org/articles?id=10.1257/aer.20101552}}

@article{abdulkadiroglu2003school,
Author = {Abdulkadiro{\u{g}}lu, Atila and Sönmez, Tayfun},
Title = {School Choice: A Mechanism Design Approach },
Journal = {American Economic Review},
Volume = {93},
Number = {3},
Year = {2003},
Month = {June},
Pages = {729–747},
DOI = {10.1257/000282803322157061},
URL = {https://www.aeaweb.org/articles?id=10.1257/000282803322157061}}

@misc{euaiact2024,
  author       = {{European Parliament and Council of the European Union}},
  title        = {Regulation ({EU}) 2024/1689 of the {E}uropean {P}arliament
                  and of the {C}ouncil of 13 {J}une 2024 laying down
                  harmonised rules on artificial intelligence
                  ({A}rtificial {I}ntelligence {A}ct)},
  howpublished = {Official Journal of the European Union, L 2024/1689},
  year         = {2024},
  note         = {12 July 2024. \url{http://data.europa.eu/eli/reg/2024/1689/oj}}
}

@misc{sffa2023,
  title        = {{Students for Fair Admissions, Inc. v. President and
                  Fellows of Harvard College}},
  howpublished = {600 U.S. 181, Supreme Court of the United States},
  year         = {2023},
  note         = {Decided June 29, 2023},
  url          ={https://www.supremecourt.gov/opinions/22pdf/20-1199_hgdj.pdf}
}

@inproceedings{singh2018fairness,
author = {Singh, Ashudeep and Joachims, Thorsten},
title = {Fairness of Exposure in Rankings},
year = {2018},
isbn = {9781450355520},
publisher = {Association for Computing Machinery},
address = {New York, NY, USA},
url = {https://doi.org/10.1145/3219819.3220088},
doi = {10.1145/3219819.3220088},
booktitle = {Proceedings of the 24th ACM SIGKDD International Conference on Knowledge Discovery \& Data Mining},
pages = {2219–2228},
numpages = {10},
location = {London, United Kingdom},
series = {KDD '18}
}

@inbook{singh2019policy,
author = {Singh, Ashudeep and Joachims, Thorsten},
title = {Policy learning for fairness in ranking},
year = {2019},
publisher = {Curran Associates Inc.},
address = {Red Hook, NY, USA},
booktitle = {Proceedings of the 33rd International Conference on Neural Information Processing Systems},
articleno = {487},
numpages = {11}
}

@article{kojima2012school,
title = {School choice: Impossibilities for affirmative action},
journal = {Games and Economic Behavior},
volume = {75},
number = {2},
pages = {685-693},
year = {2012},
issn = {0899-8256},
doi = {10.1016/j.geb.2012.03.003},
url = {https://www.sciencedirect.com/science/article/pii/S0899825612000437},
author = {Fuhito Kojima}
}

@article{hafalir2013effective,
author = {Hafalir, Isa E. and Yenmez, M. Bumin and Yildirim, Muhammed A.},
title = {Effective affirmative action in school choice},
journal = {Theoretical Economics},
volume = {8},
number = {2},
pages = {325-363},
doi = {10.3982/TE1135},
url = {https://onlinelibrary.wiley.com/doi/abs/10.3982/TE1135},
eprint = {https://onlinelibrary.wiley.com/doi/pdf/10.3982/TE1135},
year = {2013}
}

\end{document}